\documentclass[sigconf]{acmart}

\usepackage{graphicx}
\usepackage{subfigure}
\usepackage{algorithmic}
\usepackage{algorithm}
\usepackage{multirow}
\usepackage{amsthm}
\usepackage{amsmath}
\newtheorem{myDef}{Definition}

\newcommand{\eg}{\textit{e.g.}}
\newcommand{\ie}{\textit{i.e.}}

\copyrightyear{2022}
\acmYear{2022}
\setcopyright{acmlicensed}
\acmConference[WSDM '22] {Proceedings of the Fifteenth ACM International Conference on Web Search and Data Mining}{February 21--25, 2022}{Tempe, AZ, USA.}
\acmBooktitle{Proceedings of the Fifteenth ACM International Conference on Web Search and Data Mining (WSDM '22), February 21--25, 2022, Tempe, AZ, USA}
\acmPrice{15.00}
\acmISBN{978-1-4503-9132-0/22/02}
\acmDOI{10.1145/3488560.3498473}

\ccsdesc[500]{Computing methodologies~Semi-supervised learning settings}
\ccsdesc[500]{Computing methodologies~Neural networks}
\ccsdesc[300]{Computing methodologies~Anomaly detection}

\keywords{Graph-level anomaly detection, Graph neural networks, Knowledge distillation, Deep learning}

\begin{document}
\fancyhead{}

\title[Deep Graph-level Anomaly Detection by Glocal Knowledge Distillation]{Deep Graph-level Anomaly Detection\\ by Glocal Knowledge Distillation}

\author{Rongrong Ma}
\affiliation{%
  \institution{Australian Artificial Intelligence Institute \\ University of Technology Sydney}
  \city{Sydney}
  \country{Australia}}
\email{rongrong.ma-1@student.uts.edu.au}

\author{Guansong Pang}
\authornote{Corresponding author: Guansong Pang, Ling Chen.}
\affiliation{%
  \institution{School of Computing and Information Systems \\ Singapore Management University}
  \country{Singapore}}
\email{gspang@smu.edu.sg}

\author{Ling Chen}
\authornotemark[1]
\affiliation{%
  \institution{Australian Artificial Intelligence Institute \\ University of Technology Sydney}
  \city{Sydney}
  \country{Australia}}
\email{ling.chen@uts.edu.au}

\author{Anton van den Hengel}
\affiliation{%
  \institution{Australian Institute for Machine Learning\\ The University of Adelaide}
  \city{Adelaide}
  \country{Australia}}
\email{anton.vandenhengel@adelaide.edu.au}

\begin{abstract}
Graph-level anomaly detection (GAD) describes the problem of detecting graphs that are abnormal in their structure and/or the features of their nodes, as compared to other graphs.
One of the challenges in GAD is to devise graph representations that enable the detection of both \textit{locally}- and \textit{globally}-anomalous graphs,
\ie, graphs that are abnormal in their fine-grained (node-level) or holistic (graph-level) properties, respectively.
To tackle this challenge we introduce a novel deep anomaly detection approach for GAD that 
learns rich global and local normal pattern information
by \textit{joint random distillation} of graph and node representations. The random distillation is achieved by training one GNN to predict another GNN with randomly initialized network weights.  
Extensive experiments on 16 real-world graph datasets from diverse domains show that our model significantly outperforms seven state-of-the-art models. Code and datasets are available at \url{https://git.io/GLocalKD}.

\end{abstract}

\maketitle

\section{Introduction}
Graph-level anomaly detection (GAD) aims to identify graphs that are significantly different from the majority of graphs in a collection. 
The ability to record complex relationships between diverse entities renders graphs an essential and widely used representation in real-world applications. As a result, anomaly detection in graphs has broad applications in, \eg, recognizing drugs with severe side-effects \cite{lee2021descriptive}, identifying toxic molecules from chemical compound graphs \cite{aggarwal2010graph}, and breaking drug-smuggling networks \cite{varfis2011contraffic}.

\begin{figure}[t!]
  \includegraphics[width=6.5cm]{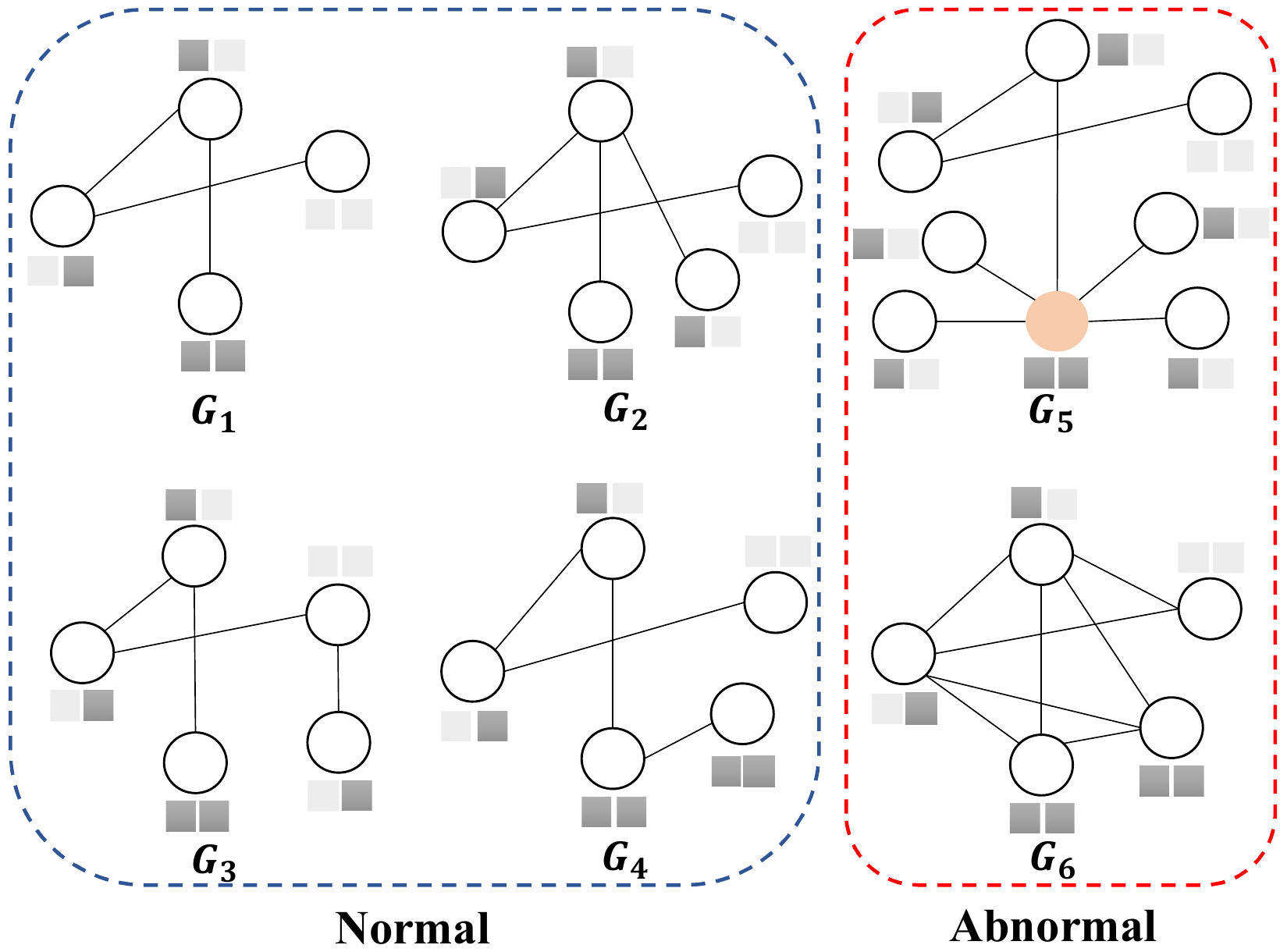}
  \centering
  \caption{A set of graphs with two anomalous graphs indicated. The squares above/below the nodes represent node features. $G_5$ is a locally-anomalous graph due to the unusual local properties (\eg, structure) of the orange node, while $G_6$ is a globally-anomalous graph because it does not conform to $G_1$ to $G_4$ in holistic graph properties.}
  \label{anomalytype}
\end{figure}

Despite the prevalence of graph data and the importance of anomaly detection therein, GAD has received little attention compared to anomaly detection in other types of data \cite{akoglu2015graph,pang2021deep}. 
One primary challenge in GAD is to learn expressive graph representations that capture local and global normal patterns in the graph structure and attributes (\eg, descriptive features of nodes). This is essential for the detection of both \textbf{locally-anomalous graph} -- relating to individual nodes and their local neighborhood ($G_5$ in Figure \ref{anomalytype}) -- and \textbf{globally-anomalous graph} -- relating to holistic graph characteristics ($G_6$ in Figure \ref{anomalytype}).

A related research line is to explore the identification of unusual changes in graph structure from a time-evolving sequence of a single graph, in which most nodes and structure at different time steps do not change~\cite{eswaran2018spotlight,sricharan2014localizing,yoon2019fast,manzoor2016fast,yu2018netwalk,zheng2019addgraph}.
GAD, in contrast, requires identifying graph anomalies among a set of graphs that lack the cohesion of a time-ordered progression and have diverse structure and node features, and it is significantly less explored.

Deep learning has shown tremendous success in diverse representation learning tasks, including
the recently emerged graph neural networks (GNN)-based methods~\cite{wu2020comprehensive}. Also, deep anomaly detection models, such as autoencoder (AE)-based methods~\cite{hawkins2002outlier,chen2017outlier,zhou2017anomaly}, generative adversarial network (GAN)-based methods~\cite{schlegl2019f,ngo2019fence} and one-class classifiers~\cite{ruff2018deep,perera2019learning,zheng2019one}, have shown promising performance on different types of data (\eg, tabular data, image data, and video data) \cite{pang2021deep}. There is limited work exploiting GNNs for the GAD task, however. A number of GNN-based models have been introduced for anomaly detection in graph data~\cite{jiang2019anomaly,ding2020inductive,kumagai2020semi,zhao2020error,wang2021one}, but they focus on anomalous node/edge detection in a single large graph.

One challenge in adapting AE- and GAN-based detection methods to GAD is their dependence on reconstruction-error-based anomaly measures. This is because it is still challenging to faithfully reconstruct (or generate) graphs from a latent vector representation~\cite{wu2020comprehensive}.
As shown by a comparative study in~\cite{zhao2020using}, the one-class model based on deep support vector data description (Deep SVDD)~\cite{ruff2018deep} may be adapted for GAD by directly optimizing the SVDD objective on top of GNN-based graph representations, but it focuses on detecting globally-anomalous graphs only. Further, its performance is largely restricted by the one-class hypersphere assumption of SVDD since there are often more complex distributions in the normal class in real-world datasets.

\begin{figure}[t!]
   \centering
   \subfigure[Random features]{\includegraphics[width=2.75cm]{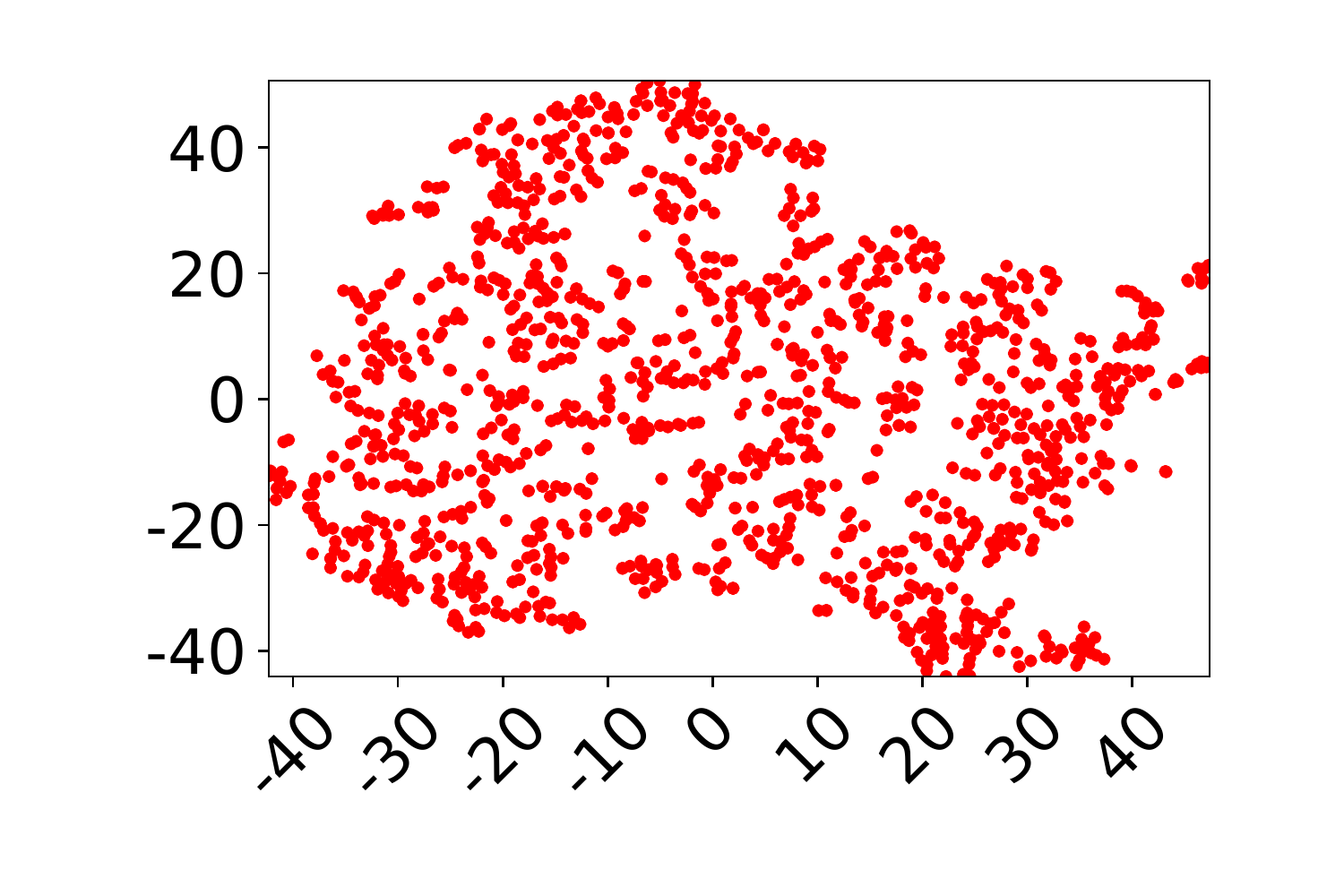}}
   \subfigure[Learned features]{\includegraphics[width=2.75cm]{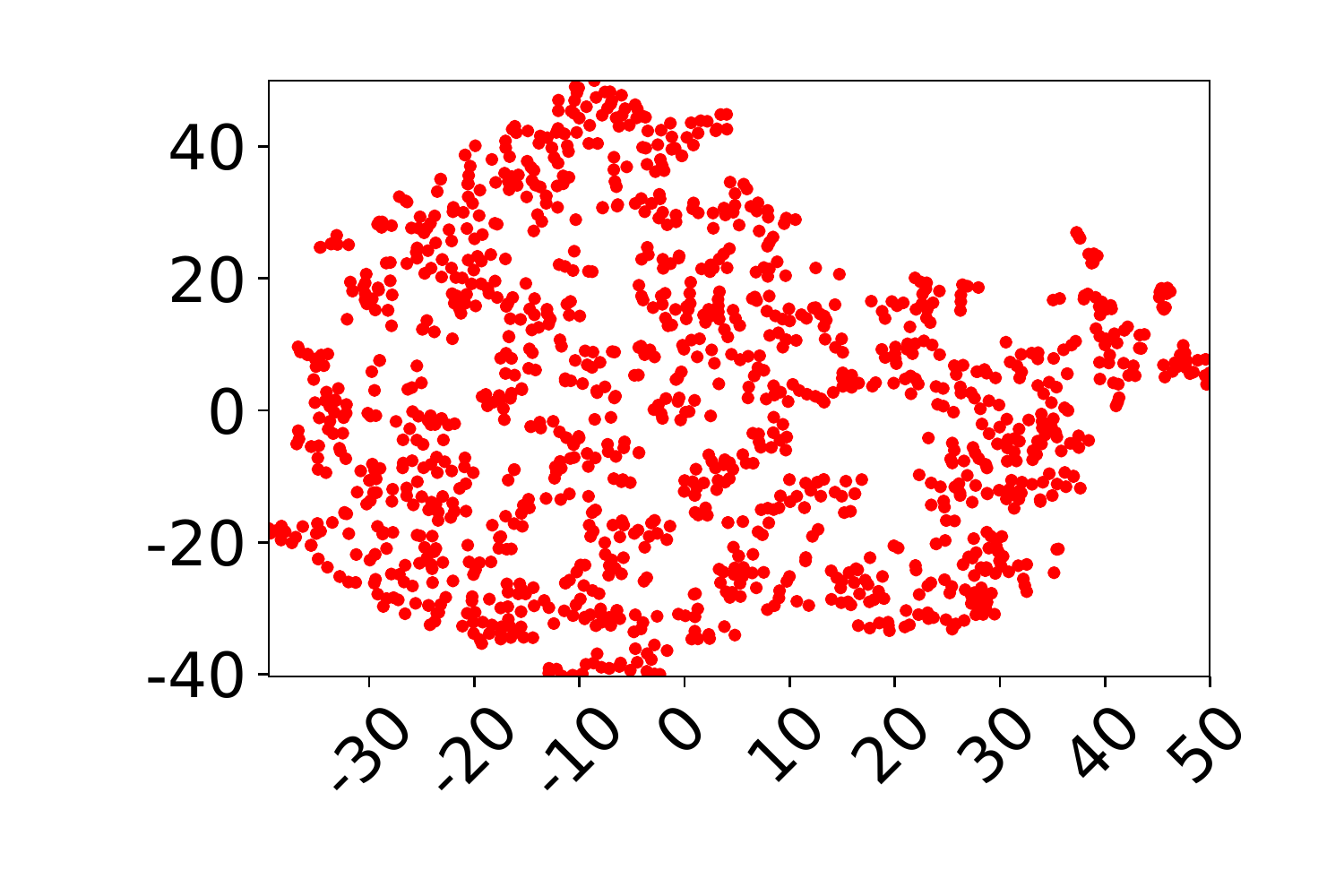}}
   \subfigure[Prediction errors]{\includegraphics[width=2.75cm]{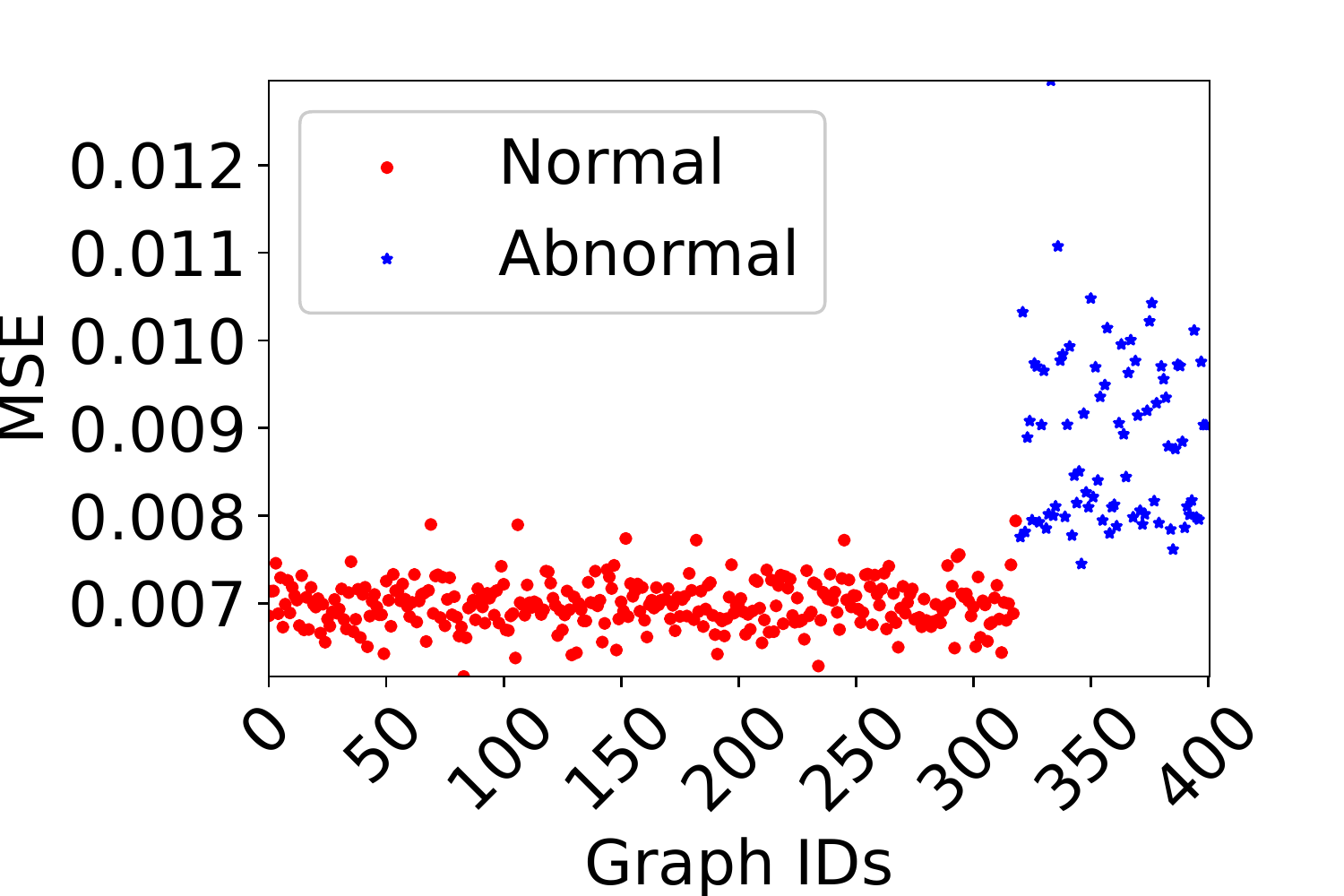}}
   \caption{Demonstration of GLocalKD working on a popular dataset -- AIDS. (a) Representations of training graphs output by the random target network. (b) Representations of training graphs learned by the predictor network. (c) Prediction errors (anomaly scores) of GLocalKD on test graphs.
    Visualization in (a) and (b) is based on t-SNE.}
   \label{aids}
\end{figure}

In this paper, we introduce a novel deep anomaly detection approach for GAD that learns both global and local normal patterns by \textbf{joint random distillation} of graph and node representations -- global and local (\ie, glocal) graph representation distillation. The random representation distillation is done by training one GNN to predict a \textbf{random GNN} that has its neural network weights fixed to random initialization, \ie, the predictor network learns to produce the same representations as that in the random network, as shown in Figure \ref{aids}(a) and (b). To accurately predict these fixed randomly-projected representations, the predictor network is enforced to learn all major patterns in the training data. By applying such a random distillation on both graph and node representations, our model learns glocal graph patterns across the given training graphs. When the training data consists of exclusively (or mostly) normal graphs, the learned patterns are a summarization of multi-scale graph regularity/normality information. As a result, given a graph that shows node/graph-level irregularity/abnormality w.r.t. these learned patterns, the model cannot accurately predict its representations, leading to a much larger prediction error than that of normal graphs, as shown in Figure \ref{aids}(c). Thus, this prediction error can be defined as anomaly score to detect the aforementioned two types of graph anomalies.

Accordingly, this paper makes the following major contributions:
\begin{itemize}
    \item We formulate the GAD problem as the task of detecting locally- or globally-anomalous graphs (Sec. \ref{subsec:problem}), and empirically verify the presence of these two types of graph anomalies in real-world datasets (Sec. \ref{subsec:ablation}).
    \item We introduce a novel deep anomaly detection framework that models \textit{glocal graph regularity} and learns graph anomaly scores in an end-to-end fashion (Sec. \ref{subsec:framework}). This results in the first approach specifically designed to effectively detect both types of anomalous graphs.
    \item A new GAD model, namely \underline{G}lobal and \underline{Local} \underline{K}nowledge \underline{D}istillation (GLocalKD), is further instantiated from the framework. GLocalKD implements the joint random distillation of graph and node representations by minimizing the graph- and node-level prediction errors of approximating a random graph convolutional neural network (Sec. \ref{sec:model}). GLocalKD is easy-to-implement without requiring the challenging graph generation, and it can effectively learn diverse glocal normal patterns with small training data. It also shows remarkable robustness to anomaly contamination, indicating its applicability in both unsupervised (anomaly-contaminated unlabeled training data) and semi-supervised (exclusively normal training data) settings.
\end{itemize}
Extensive empirical results on 16 real-world datasets from chemistry, medicine, and social network domains show that (i) GLocalKD significantly outperforms seven state-of-the-art competing methods (Sec. \ref{subsec:sota}); (ii) GLocalKD is substantially more sample-efficient than other deep detectors (Sec. \ref{subsec:data efficiency}), \eg, it can use $95\%$ less training samples to achieve the accuracy that still outperforms the competing methods by a large margin; and (iii) GLocalKD, using a single default GNN architecture, performs very stably w.r.t. different anomaly contamination rates (Sec. \ref{subsec:robustness}) and the dimensionality of the representations (Sec. \ref{subsec:sensitivity}).

\section{Related Work}
\subsection{Graph-level Anomaly Detection}
Graph-based anomaly detection has drawn great attention in recent years~\cite{akoglu2015graph}, especially the recently emerged GNN-based approaches \cite{ding2020inductive,wang2021one,kumagai2020semi,jiang2019anomaly,zhao2020error}, 
but most of the studies focus on anomaly (\eg, anomalous nodes or edges) detection in a single large graph. Below we review related work on GAD.

\noindent \textbf{Time-evolving Graphs}.
Most existing GAD studies are to identify anomalous graph changes in a sequence of time-evolving graphs~\cite{eswaran2018spotlight,sricharan2014localizing,yoon2019fast,manzoor2016fast,yu2018netwalk,zheng2019addgraph,lagraa2021simple}.
However, these methods are designed to handle time-dependent graphs with very similar structure and difficult to generalize to graphs with large variations in the structure and/or descriptive features.

\noindent \textbf{Static Graphs}. Significantly less work has been done on anomalous graph detection in a set of static graphs. One research direction is to utilize powerful graph representation methods or graph kernels for GAD. A number of recent studies \cite{nguyen2020anomaly,zhao2020using} show promising GAD performance by applying off-the-shelf anomaly measures, such as isolation forest (iForest)~\cite{liu2008isolation}, local outlier factor (LOF)~\cite{breunig2000lof}, one-class support vector machine (OCSVM)~\cite{scholkopf1999support}, on top of vectorized graph representations learned by advanced graph kernels (such as Weisfeiler-Leman kernel (WL)~\cite{shervashidze2011weisfeiler} and propagation kernel (PK)~\cite{neumann2016propagation}) or graph representation learning methods (such as Graph2Vec~\cite{narayanan2017graph2vec} and InfoGraph~\cite{sun2019infograph}). The key issue with these methods is that the graph representations are learned independently from the anomaly detectors, leading to suboptimal representations. Alternatively, there are studies on extracting graph-level patterns for GAD~\cite{nguyen2020anomaly}. However, the performance of these methods is limited because graph-level patterns may differ significantly in graphs from different domains or application scenarios.

\noindent \textbf{Deep Learning-based Methods}. 
Despite great success of deep anomaly detection in different types of data~\cite{pang2021deep}, there is limited work done on GAD in this line. Deep graph learning techniques, such as graph convolutional network (GCN)~\cite{kipf2016semi} and graph isomorphism network (GIN)~\cite{xu2018powerful}, have been powerful graph representation learning tools that empower diverse downstream tasks \cite{zhang2020deep,wu2020comprehensive}. Most of existing deep anomaly detection methods~\cite{hawkins2002outlier,chen2017outlier,zhou2017anomaly,schlegl2019f,pang2018learning,ruff2018deep,perera2019learning,zheng2019one,pang2019devnet} depend heavily on data reconstruction/generative models. Consequently, the difficulty of reconstructing/generating graphs largely hinders the development of those deep methods for GAD.
Zhao et al. \cite{zhao2020using} performs a large evaluation study on GAD, which shows that the Deep SVDD objective \cite{ruff2018deep} can be applied on top of GNN-based graph representations for enabling GAD. Nevertheless, it is focused on high-level graph anomalies only and its performance is also restricted by the SVDD measure.

\subsection{Knowledge Distillation}
Knowledge Distillation (KD), where the initial goal is to train a simple model that distills the knowledge of a large model while maintaining similar accuracy as the large model, is first introduced in \cite{hinton2015distilling} and then extended to anomaly detection in a number of studies~\cite{salehi2020multiresolution,bergmann2020uninformed,xiao2021unsupervised,li2020face,georgescu2021anomaly}. All of these methods train a simpler student network to distill the knowledge of \textbf{a pretrained teacher network on large-scale data}, such as ResNet/VGG networks pretrained on ImageNet \cite{bergmann2020uninformed,salehi2020multiresolution}. 
However, for learning tasks on graph-level data, no such general-purpose pretrained teacher networks are available; further, graph databases from different domains differ significantly from each other, which also prevents the application of this type of approach to the GAD task.
Random knowledge distillation is originally introduced in~\cite{burda2018exploration} to address sparse reward problems in deep reinforcement learning (DRL). 
It uses the random distillation errors to measure the novelty of states as some additional reward signals to encourage DRL agents' exploration in sparse-reward contexts. This idea is also used in \cite{wang2020rdp} to regularize unsupervised representation learning, enabling better anomaly detection on tabular data. Inspired by this, we devise the GLocalKD model to jointly learn globally- and locally-sensitive graph normality. To the best of our knowledge, this is the first approach designed specifically for deep graph-level anomaly detection and for detecting both types of graph anomalies.

\section{Framework}
\subsection{Problem Statement}\label{subsec:problem}
This work tackles the problem of end-to-end graph-level anomaly detection. Specifically, given a set of $M$ normal graphs $\mathcal{G}=\{G_1,...,G_M\}$, we aim at learning an anomaly scoring function $f: \mathcal{G} \rightarrow \mathbb{R}$, parameterized by $\Theta$, such that $f(\hat{G_i};\Theta) > f(\hat{G_j};\Theta)$ if $\hat{G_j}$ conforms to $\mathcal{G}$ better than $\hat{G_i}$. 
In $\mathcal{G}$, each graph is denoted by $G=(\mathcal{V}_G,\mathcal{E}_G)$ with a vertex/node set $\mathcal{V}_G$ and an edge set $\mathcal{E}_G$. The graph structure for each $G$ can be denoted by an adjacency matrix $\mathbf{A}\in\mathbb{R}^{N\times N}$ where $N$ is the number of nodes in $G$, \ie, $\mathbf{A}(i,j)=1$ if there exists an edge between nodes $v_i$ and $v_j$ ($\exists\; (v_i, v_j)\in \mathcal{E}_G$); and $\mathbf{A}(i,j)=0$ otherwise. Each node of $G$, $v_i\in \mathcal{V}_G$, is further associated with a feature vector $\mathbf{x}_i\in\mathbb{R}^n$ if $G$ is an \textit{attributed graph}. $G$ is otherwise a \textit{plain graph}. As shown in our experiments, our approach is flexible to handle both types of graph data (see Table~\ref{auroc}), and it also performs well in unsupervised settings where $\mathcal{G}$ is anomaly-contaminated and contains some unknown abnormal graphs (Sec. \ref{subsec:robustness}).

Anomalous graphs in a graph set can be classified into two categories, \ie, locally-anomalous graphs and globally-anomalous graphs, which are respectively defined as follows.

\begin{myDef}[Locally-anomalous Graph]
Given a graph data set $\mathcal{G}=\{G_i\}_i^M$, with each graph $G\in\mathcal{G}$ denoted by $G=(\mathcal{V}_G,\mathcal{E}_G)$,
graph $\hat{G}$ is a locally-anomalous graph if $\hat{G}$ does not conform to the graphs in $\mathcal{G}$ due to the presence of some anomalous nodes $v$, $\forall v \in \mathcal{V}_{\hat{G}}$, that significantly deviate from similar nodes in the graphs in $\mathcal{G}$.
\end{myDef}
\begin{myDef}[Globally-anomalous Graph]
Given a graph data set $\mathcal{G}=\{G_i\}_i^M$, graph $\hat{G}$ is a globally-anomalous graph if the holistic graph properties of $\hat{G}$ do not conform to that of the graphs in $\mathcal{G}$.
\end{myDef}

We aim to train a detection model that can detect
these two types of abnormal graphs. 
Note that the detection of locally-anomalous graphs is different from anomalous node detection in~\cite{jiang2019anomaly,ding2020inductive,kumagai2020semi,wang2021one} because the former is to detect graphs by evaluating the nodes/edges across a set of independent and separate graphs while the latter is to detect nodes/edges given a set of dependent nodes and edges from a single graph.

\subsection{The Proposed Framework}\label{subsec:framework}
To solve the above problem, we propose an end-to-end scoring framework that synthesizes two graph neural networks and joint random knowledge distillation of graph and node representations to train a deep anomaly detector. The resulting model can effectively detect both types of anomalous graphs.

\subsubsection{Overview of the Framework}
Our framework jointly distills graph-level and node-level representations of each graph, to learn both global and local graph normality information. It consists of two graph neural networks -- a fixed randomly initialized target network and a predictor network -- with exactly the same architecture and two distillation losses. It learns the holistic (fine-grained) graph normality  by training the predictor network to predict the graph (node) level representations produced by the random target network. Let $\mathbf{h}_G$ and $\hat{\mathbf{h}}_G$ respectively be the graph representation of $G$ yielded by the predictor and target networks, and $\mathbf{h}_i$ and $\hat{\mathbf{h}}_i$ be the respective node representation for a node $v_i$ in $G$ produced by the two networks, the overall objective of our approach can be given as:
\begin{equation}\label{eqn:overallobjective}
    L=L_{\mathit{graph}} +\lambda L_{\mathit{node}},
\end{equation}
where $\lambda$ is a hyperparameter that balances the importance of the two loss functions, $L_{\mathit{graph}}$ and $L_{\mathit{node}}$ are respective graph-level and node-level distillation loss functions:
\begin{align}
    L_{\mathit{graph}} &= \frac{1}{|\mathcal{G}|}\sum_{G\in \mathcal{G}}\mathit{KD}\left(\mathbf{h}_G, \hat{\mathbf{h}}_G\right),\label{eqn:graph}\\
    L_{\mathit{node}} &= \frac{1}{|\mathcal{G}|}\sum_{G\in \mathcal{G}}\left(\frac{1}{|G|}\sum_{v_i\in \mathcal{V}_G}\mathit{KD}\left(\mathbf{h}_i, \hat{\mathbf{h}}_i\right)\right),\label{eqn:node}
\end{align}
where $\mathit{KD}(\cdot,\cdot)$ is a distillation function
that measures the difference between two feature representations and $|\mathcal{G}|$ is the number of graphs in $\mathcal{G}$.

\begin{figure}[t!]
  \centering
  \includegraphics[width=8.7cm]{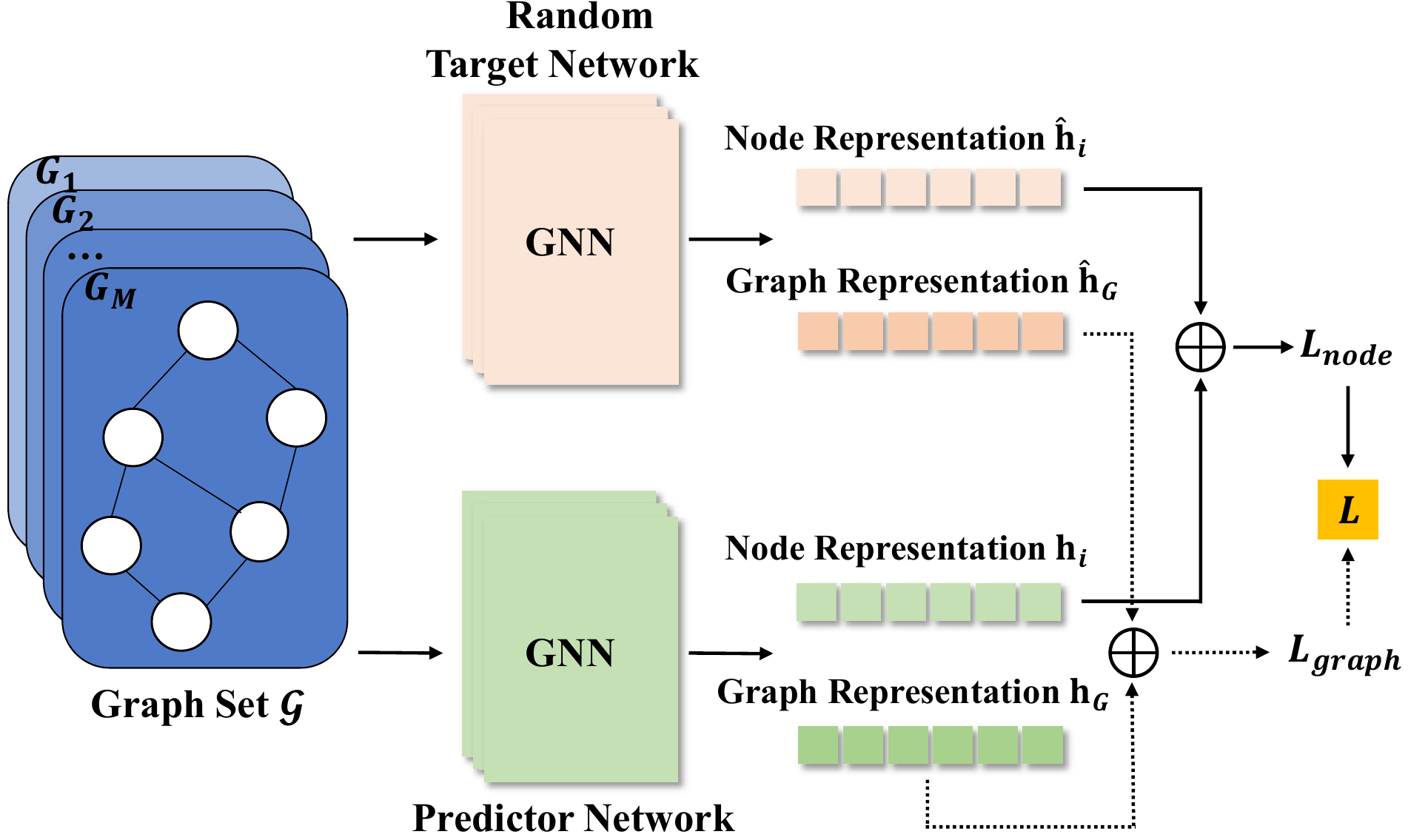}
  \caption{The proposed framework}
  \label{framework}
\end{figure}

The overall procedure of the training stage of our framework is shown in Figure \ref{framework}, which works as follows: 
\begin{itemize}
    \item [(1)] 
    We first randomly initialize a graph network $\hat{\phi}$ as the target network and fix its weight parameters $\hat{\mathbf{\Theta}}$. For every given graph $G$, it will yield a graph-level representation $\hat{\mathbf{h}}_G$ and node-level representation $\hat{\mathbf{h}}_i$ for each node $v_i$ in $G$.
    \item [(2)]
    A predictor network $\phi$, with the same architecture as $\hat{\phi}$, is parameterized by $\Theta$ and trained to predict the representation outputs of the target network $\hat{\phi}$. That is, for every given graph $G$, it produces the graph-level representation $\mathbf{h}_G$ and the node-level representation $\mathbf{h}_i$, $\forall v_i \in \mathcal{V}_G$.
    \item [(3)]
    Lastly, for graph $G$, $\hat{\mathbf{h}}_G$, $\mathbf{h}_G$, $\hat{\mathbf{h}}_i$, and $\mathbf{h}_i$ are integrated into a loss function $L$, which is minimized to train the predictor network $\phi$.
\end{itemize}

At the evaluation stage, the anomaly score for a given graph $G$ is defined as
\begin{equation}\label{eqn:framework_evaluation}
    f(G;\hat{\mathbf{\Theta}},\Theta^*) = \mathit{KD}\left(\mathbf{h}_G, \hat{\mathbf{h}}_G\right) + \lambda \frac{1}{|G|}\sum_{v_i\in \mathcal{V}_G}\mathit{KD}\left(\mathbf{h}_i, \hat{\mathbf{h}}_i\right),
\end{equation}
where $\Theta^*$ are the learned parameters of the predictor network.

\subsubsection{Key Intuition}
The graph-level and node-level representations of graphs are learned by GNNs, whose powerful capabilities of capturing graph structure and semantic information have been proved in various learning tasks and applications. The joint random distillation in our framework forces both graph representations and node representations of the predictor network to be as close as possible to the corresponding outputs of the fixed random target network on normal graph data. This resembles the extraction of different patterns (either frequently or infrequently) presented in the random representations of graphs and nodes, respectively. If a pattern frequently occurs in the random representation space, the pattern would be distilled better, \ie, the prediction error in Eq. \ref{eqn:graph} or \ref{eqn:node} is small due to a large sample size of the pattern; and the prediction error is large otherwise. As a result, our joint random distillation learns such regularity information from both graph and node representations. For a given test graph $G$, its anomaly score $f(G;\hat{\mathbf{\Theta}},\Theta^*)$ would be large if it does not conform to the regularity information embedded in the training graph set $\mathcal{G}$ at either the graph or the node level, \eg, $G_5$ and $G_6$ in Figure \ref{anomalytype}; and $f(G;\hat{\mathbf{\Theta}},\Theta^*)$ would be small otherwise, \eg, $G_1\; -\; G_4$ in Figure \ref{anomalytype}.

\section{Joint Random Distillation of Graph and Node Representations} \label{sec:model}
The proposed framework is instantiated into a method called Global and Local Knowledge Distillation (GLocalKD), in which we use widely-used graph convolutional network (GCN) to learn node and graph representations and the joint distillation is driven by two mean square error-based loss functions.

\subsection{Graph Neural Network Architecture}

\subsubsection{Random Target Network}
We first establish a target network with randomly initialized weights to 
obtain graph- and node-level representations in the random space. Different graph representation approaches may be used to generate the required representations as the prediction targets of the predictor network. Theoretically, various deep graph networks, such as GCN, GAT and GIN, can be employed as the graph representation learning module. In our work, a standard GCN is used, because GCN and its variants have proved their power to learn expressive features of graphs and good computational efficiency~\cite{zhang2020deep,wu2020comprehensive}. 

Specifically, $\hat{\phi}(\cdot,\hat{\Theta}): G=(\mathcal{V}_G,\mathcal{E}_G) \rightarrow \mathbb{R}^{N\times k}$ is a GCN with fixed randomly initialized weights $\hat{\Theta}$ (\ie, the GCN is frozen after random weight initialization), where $N$ is the number of nodes in $G$ and $k$ is a predefined dimensionality size of node representations. For each graph $G=(\mathcal{V}_G,\mathcal{E}_G)$ in $\mathcal{G}$, $\hat{\phi}(\cdot)$ takes adjacency matrix $\mathbf{A}$ and feature matrix $\mathbf{X}$ as input, and maps each node $v_i\in \mathcal{V}_G$ to the representation space using $\hat{\Theta}$.
Let $\hat{\mathbf{h}}_i^l$ be the hidden representation of node $v_i$ in the $l^{th}$ layer, which is formally computed as follows: 
\begin{equation}
    \hat{\mathbf{h}}_i^l=\rho\left(\sum_{j\in\widetilde{\mathcal{N}}(i)}\frac{1}{\sqrt{\widetilde{\mathbf{D}}(i,i)\widetilde{\mathbf{\mathbf{D}}}(j,j)}}\hat{\mathbf{h}}_j^{l-1}\hat{\Theta}^{l-1} \right)
\end{equation}
where 
$\hat{\mathbf{h}}_j^{l-1}$ represents the hidden representation of node $v_j$ in the $(l-1)^{th}$ layer, $\rho(a)=max(0,a)$ is the ReLU activation function, $\mathcal{N}(i)$ denotes the 1st-order neighbors of $v_i$ and $\widetilde{\mathcal{N}}(i)=\mathcal{N}(i)\cup\{v_i\}$, $\mathbf{D}$ is a diagonal degree matrix with $\mathbf{D}(i,i)=\sum_j\mathbf{A}(i,j)$, $\widetilde{\mathbf{\mathbf{D}}}=\mathbf{D}+\mathbf{I}$ ($\mathbf{I}$ is an identity matrix), and the input representation of $v_i$ in the $0^{th}$ layer, $\hat{\mathbf{h}}_i^0$, is initialized by its feature vector in $\mathbf{X}$, \ie, $\hat{\mathbf{h}}_i^0=\mathbf{X}(i,:)$.
Thus, the output random node representation $\hat{\mathbf{h}}_i$ for node $v_i$ can be written as:
\begin{equation}\label{targetnode}
    \hat{\mathbf{h}}_i = \rho\left(\sum_{j\in\widetilde{\mathcal{N}}(i)}\frac{1}{\sqrt{\widetilde{\mathbf{D}}(i,i)\widetilde{\mathbf{\mathbf{D}}}(j,j)}}\hat{\mathbf{h}}_j^{K-1}\hat{\Theta}^{K-1} \right)
\end{equation}
where $K$ is the number of layers of $\hat{\phi}(\cdot)$. 
The feature matrix $\mathbf{X}$ is composed of node attributes for attributed graphs. For plain graphs, following \cite{zhang2018end}, we use the node degree as the node feature to construct a simple $\mathbf{X}$, since the degree of nodes is one of the key information for the discriminability of nodes and graphs.

Next, a READOUT operation is applied to the node representations to obtain the graph-level representation for $G$. There have been a number of READOUT operations introduced, \eg, maxing, averaging, summation and concatenation~\cite{zhang2020deep,wu2020comprehensive}. Considering that our goal is to detect anomalies, we need to aggregate extreme features across the node representations.
Thus, the max-pooling is employed in the READOUT operation:
\begin{equation}\label{targetgraph}
   \mathbf{\hat{h}}_G= \max{\{\hat{\mathbf{h}}_i,\forall v_i \in \mathcal{V}_G\}}.
\end{equation}

\subsubsection{Predictor Network}
The predictor network is a graph network used to predict the output representations of the target network, $\hat{\mathbf{h}}_i$ and $\mathbf{\hat{h}}_G$. We employ a GCN with the exactly same structure as the target network as the predictor network, which is denoted as $\phi(\cdot,\mathbf{\Theta}): G=(\mathcal{V}_G,\mathcal{E}_G) \rightarrow \mathbb{R}^{N\times k}$ with the weight parameters $\mathbf{\Theta}$ to be learned. Then, similar to $\hat{\phi}$,
$\phi(\cdot,\mathbf{\Theta})$ yields the node representation $\mathbf{h}_i$ for node $v_i$ by the following formulation:
\begin{equation}\label{prenode}
    \mathbf{h}_i = \rho\left(\sum_{j\in\widetilde{\mathcal{N}}(i)}\frac{1}{\sqrt{\widetilde{\mathbf{D}}(i,i)\widetilde{\mathbf{\mathbf{D}}}(j,j)}}\mathbf{h}_j^{K-1}\Theta^{K-1} \right)
\end{equation}
After the same READOUT operation as in $\hat{\phi}$, the graph representation $\mathbf{h}_G$ is computed as follows:
\begin{equation}\label{pregraph}
    \mathbf{h}_G= \max{\{\mathbf{h}_i,\forall v_i \in \mathcal{V}_G\}}.
\end{equation}

Thus, the only difference between the random target network $\hat{\phi}(\cdot,\hat{\Theta})$ and the predictor network $\phi(\cdot,\mathbf{\Theta})$ is that $\hat{\Theta}$ is fixed after random initialization while $\mathbf{\Theta}$ needs to be learned through the following glocal knowledge distillation.

\subsection{Glocal Regularity Distillation}
We further perform glocal regularity distillation by minimizing the distance between the (graph- and node-level) representations produced by the predictor network and the target network. Specifically, the graph-level and node-level distillation loss are defined as:
\begin{equation}\label{eqn:globaldistillation}
    L_{\mathit{graph}} = \frac{1}{|\mathcal{G}|}\sum_{G\in \mathcal{G}}\|\mathbf{h}_G - \hat{\mathbf{h}}_G \|^2,
\end{equation}
\begin{equation}\label{eqn:localdistillation}
    L_{\mathit{node}} = \frac{1}{|\mathcal{G}|}\sum_{G\in \mathcal{G}}\left(\frac{1}{|G|}\sum_{v_i\in \mathcal{V}_G}\|\mathbf{h}_i- \hat{\mathbf{h}}_i\|^2\right).
\end{equation}

To learn the global and local graph regularity information simultaneously, our model is optimized by jointly minimizing the above two losses:
\begin{equation}\label{loss}
    L = L_{\mathit{graph}} + L_{\mathit{node}}.
\end{equation}
% where 
That is, $\lambda$ in Eq. \ref{eqn:overallobjective} is set to one in Eq. \ref{loss} since it is believed that it is equivalently important to detect both of locally- and globally-anomalous graphs. We will discuss in Sec. \ref{subsec:theory} in more details about why our model can learn the global and local graph regularity.

\subsection{Anomaly Detection of Using GLocalKD}
By joint global and local random distillation, the learned representations in our predictor network capture the regularity information at both the graph and node levels.
Specifically, given a test graph sample $G$, its anomaly score is defined by the prediction errors in both graph and node-level representations:
\begin{equation}
  f(G;\hat{\mathbf{\Theta}},\Theta^*) =\left\|\mathbf{h}_G-\hat{\mathbf{h}}_G\right\|^2 + \frac{1}{|G|}\sum_{v_i\in \mathcal{V}_G}\|\mathbf{h}_i- \hat{\mathbf{h}}_i\|^2.
\end{equation}

This indicates that the locally- and globally-anomalous graph anomalies are treated equally important in our anomaly scoring, sharing the same spirit as the overall objective in Eq. \ref{loss}.

\begin{table*}[h]
\caption{AUC results (mean$\pm$std) on 16 real-world graph datasets. \# Graphs: the number of graphs, $\#$ Nodes and $\#$ Edges: the average number of nodes and edges in each graph. The best performance is boldfaced.}
\label{auroc}
\setlength{\tabcolsep}{0.6mm}
\scalebox{0.9}{
\begin{tabular}{lccc|ccccccccc}
\hline
\multirow{2}*{\textbf{Dataset}} & \multirow{2}*{\textbf{\# Graphs}} & \multirow{2}*{\textbf{\# Nodes}} & \multirow{2}*{\textbf{\# Edges}} &\multicolumn{2}{c}{\textbf{InfoGraph}} & \multicolumn{2}{c}{\textbf{WL}} & \multicolumn{2}{c}{\textbf{PK}} & \multirow{2}*{\textbf{OCGCN}} & \multirow{2}*{\textbf{GLocalKD}} \\ \cline{5-10}
 & & & & \textbf{iForest} & \textbf{LESINN} & \textbf{iForest} & \textbf{LESINN} &  \textbf{iForest} & \textbf{LESINN} & &  \\
\hline
PROTEINS$\_$full & 1,113 & 39.06 & 72.82  & 0.464$\pm$0.019 & 0.336$\pm$0.047 & 0.639$\pm$0.018 & 0.712$\pm$0.053 & 0.627$\pm$0.009 & 0.572$\pm$0.031 & 0.718$\pm$0.036 & \textbf{0.785}$\pm$0.034 \\
ENZYMES & 600 & 32.63 & 62.14 &  0.483$\pm$0.027 & 0.528$\pm$0.046 & 0.498$\pm$0.029 & 0.624$\pm$0.050& 0.493$\pm$0.013 & 0.608$\pm$0.033 & 0.613$\pm$0.087 & \textbf{0.636}$\pm$0.061 \\
AIDS & 2,000 & 15.69 & 16.2 & 0.703$\pm$0.036 & 0.955$\pm$0.023 & 0.632$\pm$0.050 & 0.584$\pm$0.016 & 0.476$\pm$0.014 & 0.421$\pm$0.010 & 0.664$\pm$0.080 & \textbf{0.992}$\pm$0.004 \\
DHFR & 467 & 42.43 & 44.54 & 0.489$\pm$0.015 & \textbf{0.625}$\pm$0.028 & 0.466$\pm$0.013 & 0.596$\pm$0.056 & 0.467$\pm$0.013 & 0.568$\pm$0.054 & 0.495$\pm$0.080 & 0.558$\pm$0.030 \\
BZR & 405 & 35.75 & 38.36 & 0.528$\pm$0.060 & 0.731$\pm$0.071 & 0.533$\pm$0.032 & 0.720$\pm$0.032 & 0.525$\pm$0.052 & \textbf{0.775}$\pm$0.063 & 0.658$\pm$0.071 & 0.679$\pm$0.065 \\
COX2 & 467 & 41.22 & 43.45 & 0.580$\pm$0.052 & 0.670$\pm$0.079 & 0.532$\pm$0.027 & 0.590$\pm$0.056 & 0.515$\pm$0.036 & \textbf{0.671}$\pm$0.039 &  0.628$\pm$0.072 & 0.589$\pm$0.045 \\
DD & 1,178 & 284.32 & 715.66 & 0.475$\pm$0.012 & 0.310$\pm$0.034 & 0.699$\pm$0.006 & 0.638$\pm$0.045 & 0.706$\pm$0.010 & \textbf{0.833}$\pm$0.023 &  0.605$\pm$0.086 & 0.805$\pm$0.017 \\
NCI1 & 4,110 & 29.87 & 32.3 & 0.494$\pm$0.009 & 0.598$\pm$0.035 & 0.545$\pm$0.008 & \textbf{0.743}$\pm$0.015 & 0.532$\pm$0.006 & 0.670$\pm$0.012 &  0.627$\pm$0.015 & 0.683$\pm$0.015\\
IMDB& 1,000 & 19.77 & 96.53  & 0.520$\pm$0.028 & 0.565$\pm$0.017 & 0.442$\pm$0.032 & \textbf{0.612}$\pm$0.046 & 0.442$\pm$0.035 & 0.585$\pm$0.047 &  0.536$\pm$0.148 & 0.514$\pm$0.039 \\
REDDIT & 2,000 & 429.63 & 497.75 & 0.457$\pm$0.003 & 0.262$\pm$0.027 & 0.450$\pm$0.013 & 0.239$\pm$0.028 & 0.450$\pm$0.012 & 0.487$\pm$0.013 & 0.759$\pm$0.056 &  \textbf{0.782}$\pm$0.016 \\
HSE & 8,417 & 16.89 & 17.23 & 0.484$\pm$0.026 & \textbf{0.657}$\pm$0.051 & 0.477$\pm$0.000 & 0.528$\pm$0.000 & 0.489$\pm$0.003 & 0.469$\pm$0.016 & 0.388$\pm$0.041 &  0.591$\pm$0.001 \\
MMP & 7,558  & 17.62 & 17.98 & 0.539$\pm$0.022 & 0.571$\pm$0.037 & 0.475$\pm$0.000 & 0.307$\pm$0.000 & 0.488$\pm$0.002 & 0.322$\pm$0.008 &  0.457$\pm$0.038 &  \textbf{0.676}$\pm$0.001 \\
p53 & 8,903 & 17.92 & 18.34  & 0.511$\pm$0.014 & 0.520$\pm$0.025 & 0.473$\pm$0.000 & 0.390$\pm$0.000 & 0.486$\pm$0.004 & 0.329$\pm$0.001 &  0.483$\pm$0.017 & \textbf{0.639}$\pm$0.002 \\
PPAR-gamma & 8,451 & 17.38 & 17.72 & 0.521$\pm$0.023 & 0.541$\pm$0.036 &  0.510$\pm$0.000 & 0.461$\pm$0.000 & 0.499$\pm$0.017 & 0.388$\pm$0.015 & 0.431$\pm$0.043 & \textbf{0.644}$\pm$0.001 \\
COLLAB & 5,000 & 74.49 & 2,457.78 & 0.453$\pm$0.003 & 0.319$\pm$0.033 & 0.506$\pm$0.020 & 0.536$\pm$0.014 & 0.529$\pm$0.023 & \textbf{0.550}$\pm$0.043 &0.401$\pm$0.183 & 0.525$\pm$0.014 \\
hERG & 655 & 26.48 & 28.79 & 0.607$\pm$0.033 & 0.701$\pm$0.048 & 0.665$\pm$0.042 & \textbf{0.802}$\pm$0.047 & 0.679$\pm$0.034 & 0.798$\pm$0.052 & 0.569$\pm$0.049 & 0.704$\pm$0.049 \\
\hline
& & & p-value & 0.0005 & 0.0262 & 0.0004 &0.1089 & 0.0005 & 0.1337 & 0.0018 & $-$ \\
\hline
\end{tabular}
}
\end{table*}

\subsection{Theoretical Analysis of GLocalKD}\label{subsec:theory}
We show below that GLocalKD can normally produce a larger anomaly score for an abnormal graph than that for a normal one. Specifically, consider a regression problem with data distribution $\hat{\mathcal{G}}=\{G_i,y_i\}_i$ ($y_i$ is the regression target) and a Bayesian setting in which a prior $p(\mathbf{\Theta}^\star)$ over the parameters of a GCN $\phi(\cdot,\mathbf{\Theta}^\star)$ is considered. The aim is to calculate the posterior after iteratively updating on the data. According to \cite{burda2018exploration}, our task can then be formulated as the optimization problem below:
\begin{equation}\label{analysis}
    \min_{\mathbf{\Theta}} \mathrm{E}_{(G_i,y_i)\sim\hat{\mathcal{G}}}\|\phi(G_i,\mathbf{\Theta})+ \phi(G_i,\mathbf{\Theta}^\star)-y_i\|^2+\mathcal{R}(\mathbf{\Theta}),
\end{equation}
where $\mathcal{R}(\mathbf{\Theta})$ is a regularization term from the prior \cite{osband2018randomized}. Let $\mathcal{F}$ be the distribution over functions $f_{\mathbf{\Theta}} = \phi(\cdot,\mathbf{\Theta}) + \phi(\cdot,\mathbf{\Theta}^\star)$, where $\mathbf{\Theta}$ is the solution of Eq.~\ref{analysis} and $\mathbf{\Theta}^\star$ is drawn from $p(\mathbf{\Theta}^\star)$, then the ensemble $\mathcal{F}$ can bee seen as an approximation of the posterior \cite{osband2018randomized}.

When we select the graphs from the same distribution and set the label $y_i$ to zero, the optimization problem 
\begin{equation}
\arg \min_{\mathbf{\Theta}} \mathrm{E}_{(G_i,y_i)\sim\hat{\mathcal{G}}}\|\phi(G_i,\mathbf{\Theta}) + \phi(G_i,\mathbf{\Theta}^\star)\|^2   
\end{equation}
is equivalent to distilling a randomly drawn function from the prior. From this perspective, each entry of the representation outputs of the target and the predictor networks would correspond to a part of an ensemble and the prediction error would be an estimate of the predictive variance of the ensemble when the ensemble is assumed to be unbiased, as discussed in \cite{burda2018exploration}. If we consider $\phi(\cdot,\mathbf{\Theta}^\star)$ as the target network with randomly initialized $\mathbf{\Theta}^\star$ and regard $\phi(\cdot,\mathbf{\Theta})$ as the predictor network, the prediction errors of the node representations as well as graph representations in the predictor network would be an estimate of the predictive variance of the results of two networks. In other words, our training process aims to train a predictor network so that the node representations and graph representations of the two networks on each training sample are as close as possible.
Then, for the graph with patterns similar to many other training graphs, the prediction errors in Eqs. \ref{eqn:globaldistillation} and \ref{eqn:localdistillation} are small, \ie, small predictive variance in Eq. \ref{analysis}, because there are sufficient such samples to train the prediction model; the abnormal graphs, by contrast, are drawn from different distributions from the training graphs and dissimilar to most of the training data, leading to large predictive variance in Eq. \ref{analysis}. Thus, the prediction errors in our joint random distillation can distinguish both locally- and globally-anomalous graphs from normal graphs.

\section{Experiments and Results}
\subsection{Datasets}
As shown in Table~\ref{auroc},
% in our experiment, 
we employ sixteen publicly available real-world datasets\footnote{All of the datasets were accessed via http://graphkernels.cs.tu-dortmund.de, except hERG that was from https://tdcommons.ai/. }, which are collected from various critical domains~\cite{KKMMN2016}. 
The first six datasets in Table~\ref{auroc} are attributed graphs, \ie, each node has some descriptive features; the others are plain graphs. HSE, MMP, p53 and PPAR-gamma are datasets with real anomalies. The other 12 datasets are taken from graph classification benchmarks and converted for anomaly detection tasks by treating the minority class as anomalies, following \cite{liu2008isolation,pang2019devnet,campos2016evaluation}. 
These datasets are selected mainly because the graph samples in the chosen anomaly class meet some key semantics of anomalies, \eg, graphs are scatteredly or sparsely distributed in the representation space.

\subsection{Competing Methods}
Seven competing methods from two types of approach are used. 

\noindent \textbf{Two-step Methods.} This approach first uses state-of-the-art graph representation-based methods to obtain vectorized graph representations, and then applies advanced off-the-shelf shallow anomaly detectors on top of the representations to calculate anomaly scores. InfoGraph \cite{sun2019infograph}, WL \cite{shervashidze2011weisfeiler} and PK graph kernels \cite{neumann2016propagation} are used in our experiments. Anomaly detectors, including iForest \cite{liu2008isolation} and $k$NN ensemble (LESINN) \cite{pang2015lesinn}, are utilized. The combination of these embedding methods and detectors leads to six two-step methods. 

\noindent \textbf{End-to-end Methods}. We also compare with the one-class GCN-based method, namely OCGCN \cite{zhao2020using}, which can be trained in an end-to-end manner as GLocalKD. OCGCN is optimized using a SVDD objective on top of GCN-based representation learning.

\subsection{Implementation and Evaluation}
The target network and the predictor network in GLocalKD
share the same network architecture -- a network with three GCN layers. The dimension of the hidden layer is 512 and the output layer has 256 neural units. The learning rate is selected through the grid search, varying from $10^{-1}$ to $10^{-5}$. The batch size is 300 for all data sets except the four largest datasets HSE, MMP, p53 and PPAR-gamma, for which the batch size is 2000. 
For the competing methods, the network architecture and the optimization of OCGCN is the same as our model. The other methods are taken from their authors. We probed a wide range of hyperparameter settings in both iForest and LESINN. We found that the performance of iForest does not change much with varying hyperparameter settings, while LESINN can obtain large improvement of using one subsampling size setting over the others (see Table \ref{lesinn} in Appendix \ref{LESINN}). Due to these observations, iForest with subsampling size and the number of trees respectively set to 256 and 100 is used by default, while LESINN with the subsampling size setting that performs best on most of the datasets is used.
More implementation details can be found in Appendix \ref{subsec:implementation}.

In terms of evaluation, we use the popular anomaly detection evaluation metric -- Area Under Receiver Operating Characteristic Curve (AUC). Higher AUC indicates better performance. We report the mean AUC and standard deviation based on 5-fold cross-validation for all datasets, except HSE, MMP, p53 and PPAR-gamma that have widely-used training and test splits. For these four datasets, the results are based on five runs with different random seeds.

\subsection{Comparison to State-of-the-art Methods}\label{subsec:sota}
The AUC results of GLocalKD and its seven competing methods are reported in Table~\ref{auroc}.
Our GLocalKD model is the best performer on 7 datasets, achieving improvement ranging from 1\% to 12\% on many of these datasets compared to the best contenders per dataset, \eg, AIDS (3.7\%), PROTEINS\_full (6.7\%), PPAR-gamma (10.3\%), MMP (10.5\%), p53 (11.9\%);
and its performance is very close to the best contenders on some other datasets, such as DD and COLLAB. The consistent superiority of GLocalKD is mainly due to its capability in learning both global and local graph regularity. Its performance may drop significantly, \eg, decrease to performance equivalent to a random detector, if only one of these patterns is captured (see Table \ref{term}). The seven competing methods fail to work in many datasets mainly because their graph representations capture only partial local/global pattern information. 

We also perform a paired Wilcoxon signed rank test~\cite{woolson2007wilcoxon} to examine the significance of GLocalKD against each of the competing methods across the 16 datasets. As shown by the p-values in Table \ref{auroc}, GLocalKD significantly outperforms the iForest-based methods and OCGCN at the 99\% confidence level. The confidence level of the superiority of GLocalKD over LESINN-based methods ranges from 85\% and 95\%.
However, note that LESINN heavily relies on its subsample size (see Table \ref{lesinn} for the full results of InfoGraph-LESINN, WL-LESINN and PK-LESINN in Appendix \ref{LESINN}). GLocalKD works less effectively on COLLAB than some contenders, which may be due to the inseparability of anomalies from the normal graphs as the contenders also do not perform well on it.

In terms of computational efficiency, as shown by the results in Table \ref{timetable}
in Appendix \ref{subsectime}, GLocalKD and OCGCN have a similar time complexity and run much faster than the other methods in online detection, since iForest/LESINN methods require extra steps on top of the graph representations to compute the anomaly scores. On the other hand, GLocalKD and OCGCN are generally more computationally costly than the WL and PK based methods because GLocalKD and OCGCN typically require multiple iterations to perform well.

\subsection{Sample Efficiency}\label{subsec:data efficiency}
\subsubsection{Experiment Settings} 
This section examines the performance of our model w.r.t. the amount of training data, \ie, sample efficiency, using the deep competing method OCGCN as baseline. 
We use respective $5\%$, $25\%$, $50\%$, $75\%$ and $100\%$ of original training samples to train the models, and evaluate the performance on the same test data set. We report the results on the attributed graph datasets only. Similar results can be found on the other datasets.
\subsubsection{Findings}
\begin{figure}[t!]
  \centering
  \subfigcapskip=-4pt
  \subfigbottomskip=-6pt
  \subfigure[PROTEINS$\_$full]{\includegraphics[width=4.25cm]{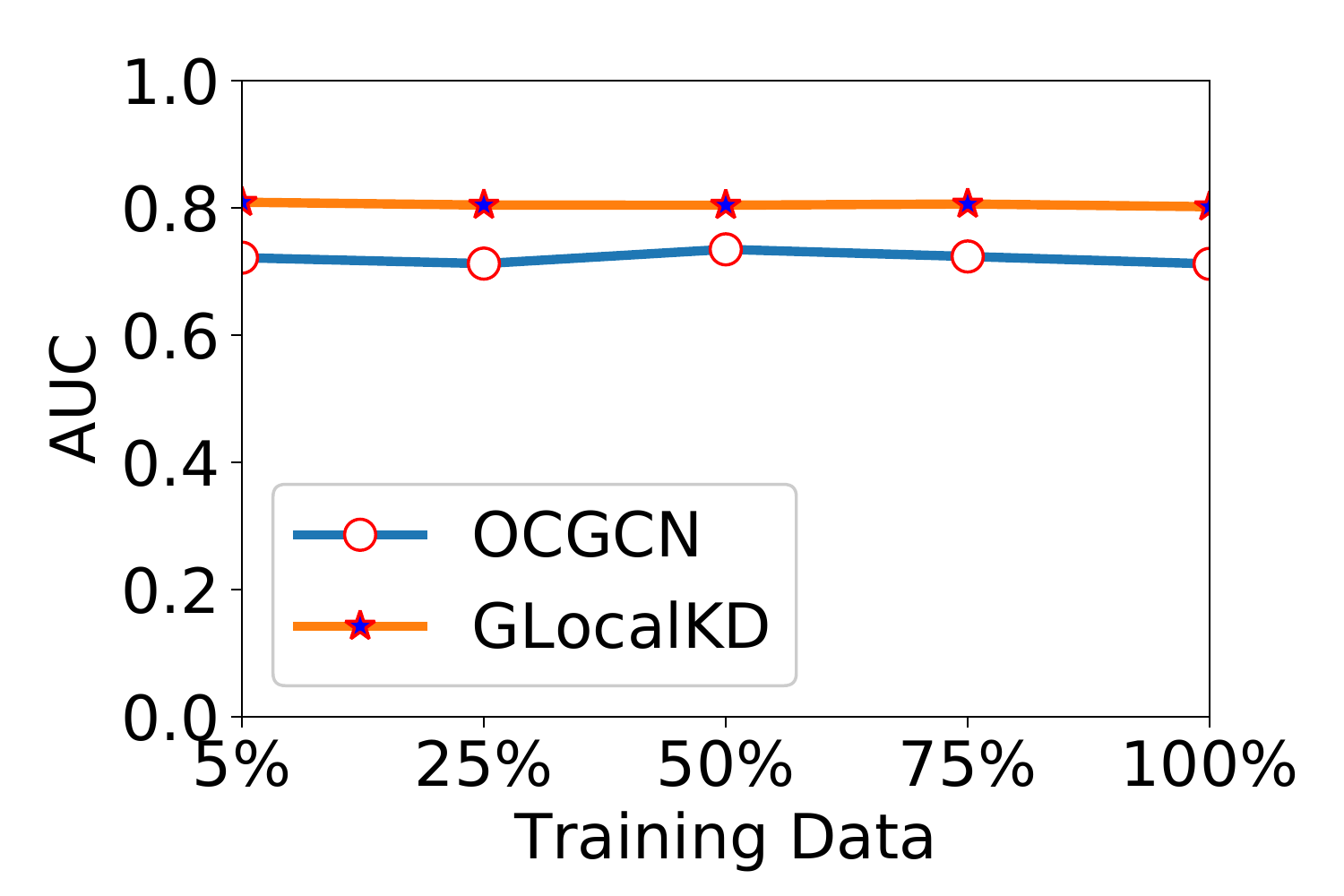}}\hspace{-6pt}
  \subfigure[ENZYMES]{\includegraphics[width=4.25cm]{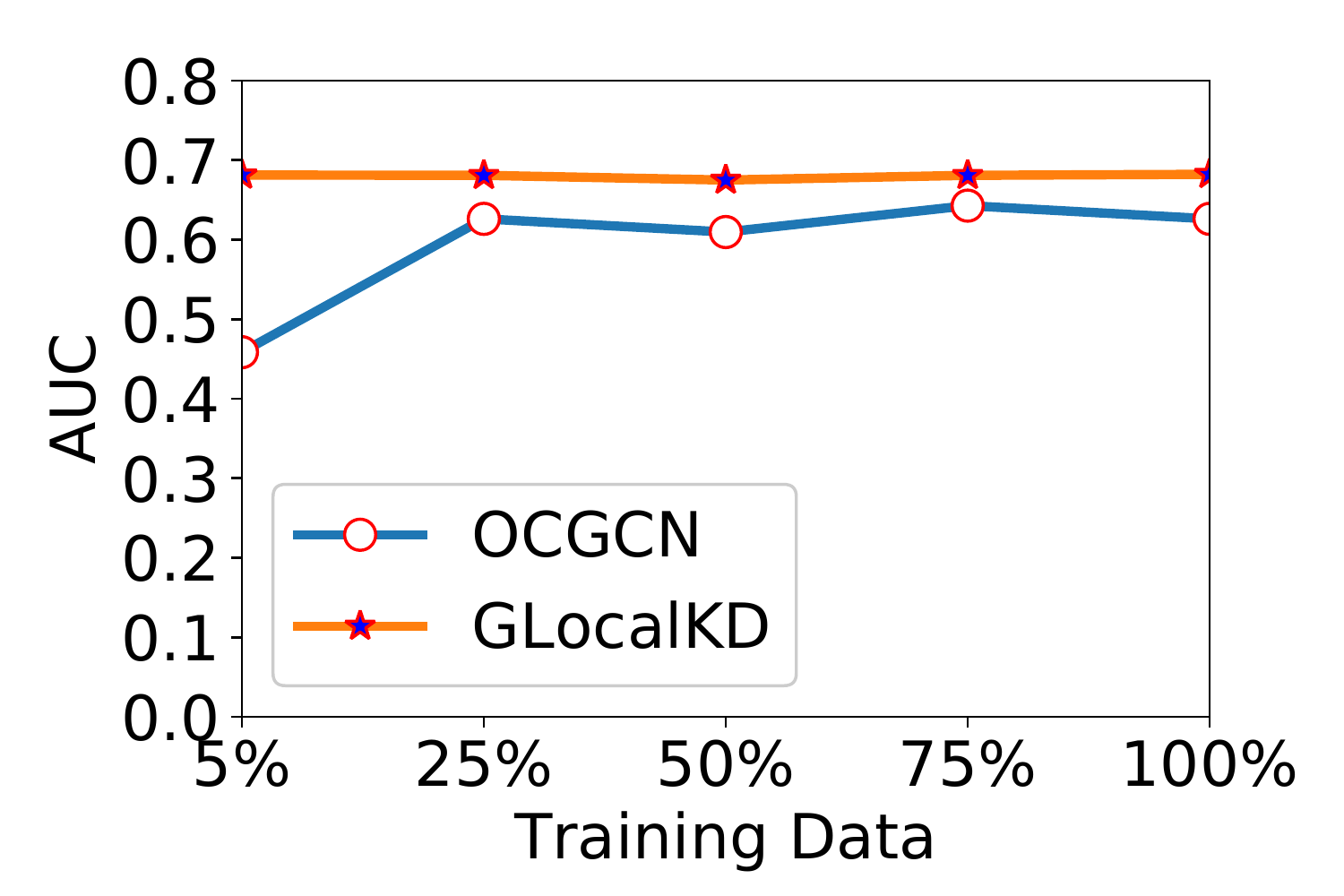}}
  \subfigure[AIDS]{\includegraphics[width=4.25cm]{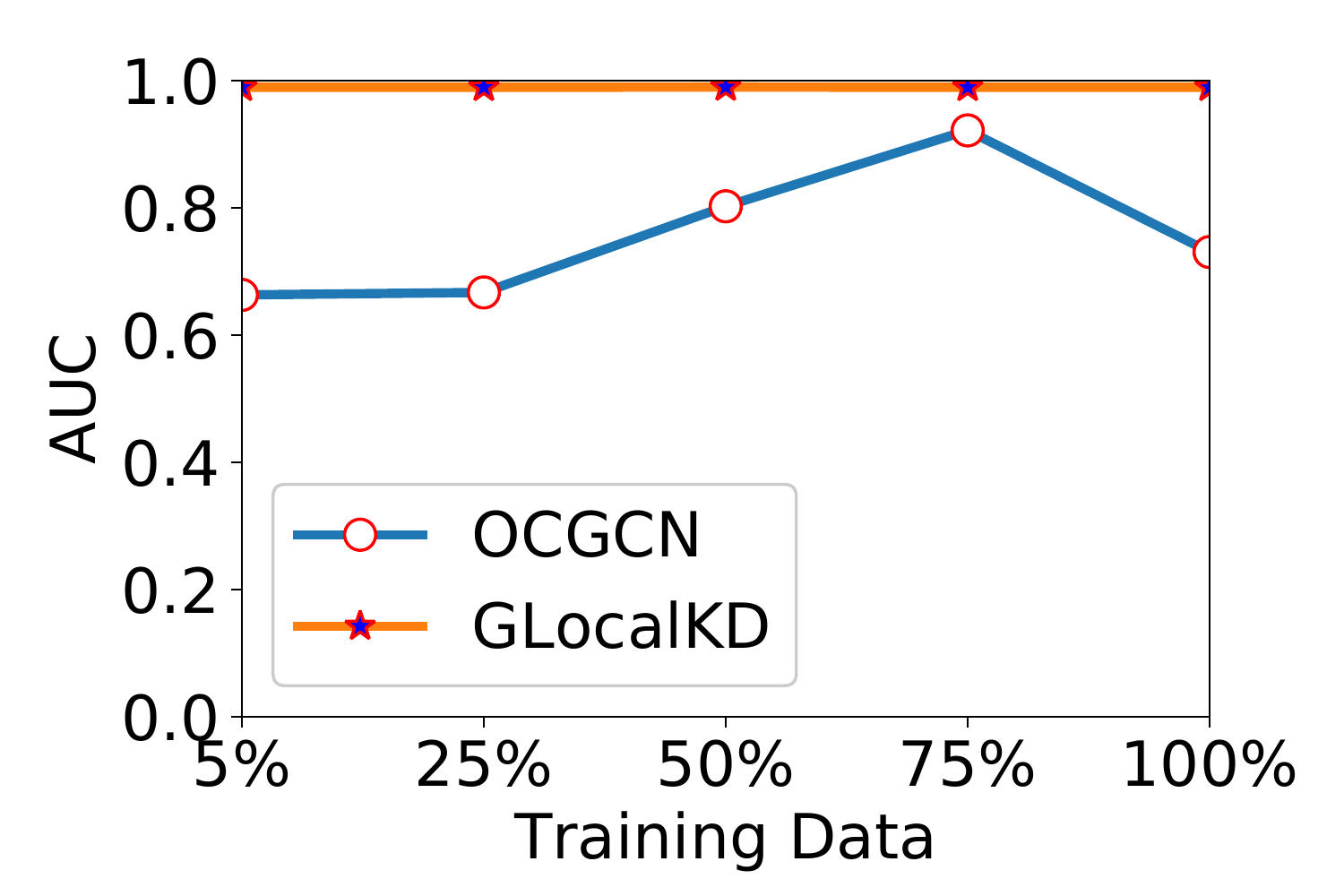}}\hspace{-6pt}
  \subfigure[DHFR]{\includegraphics[width=4.25cm]{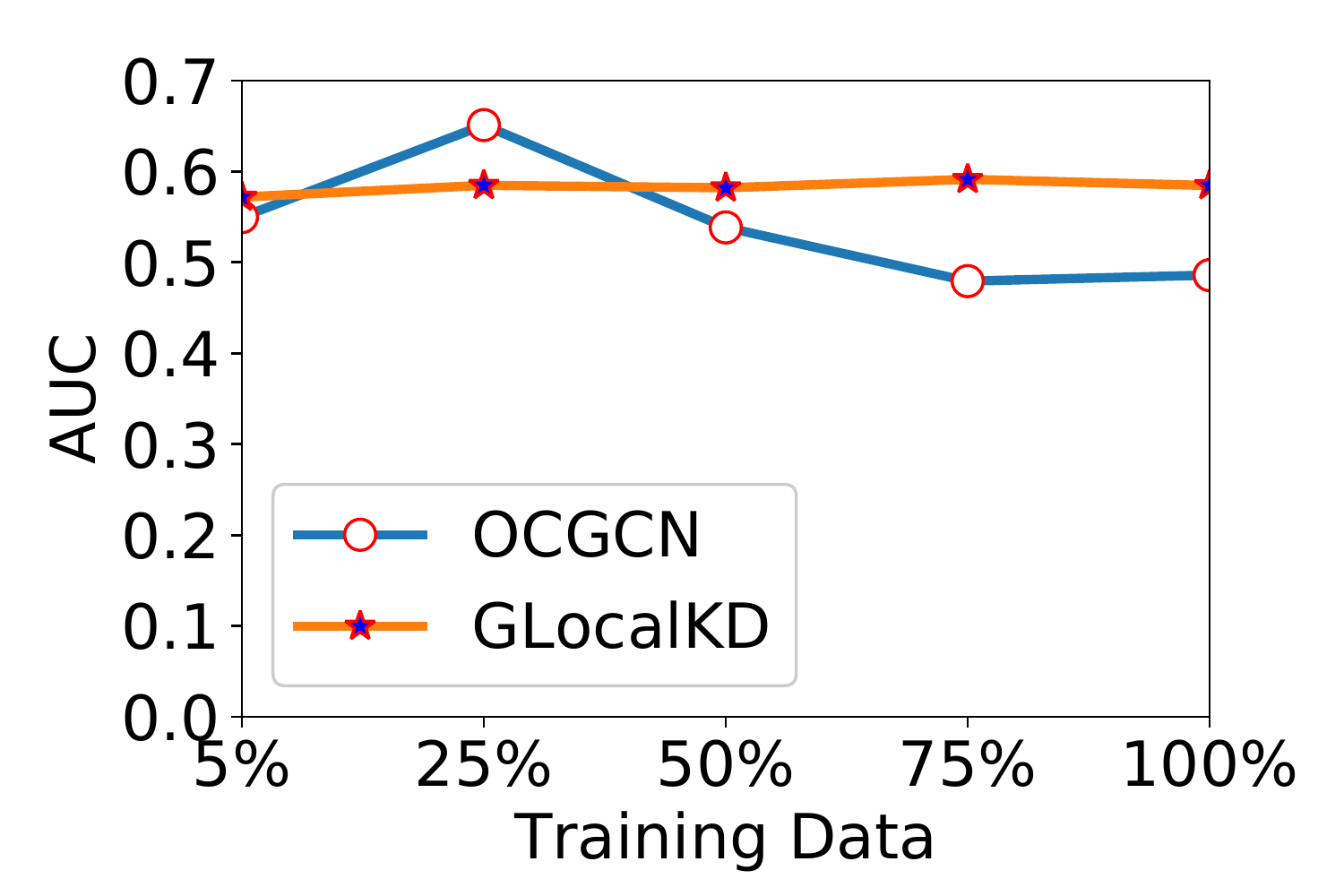}}
  \subfigure[BZR]{\includegraphics[width=4.25cm]{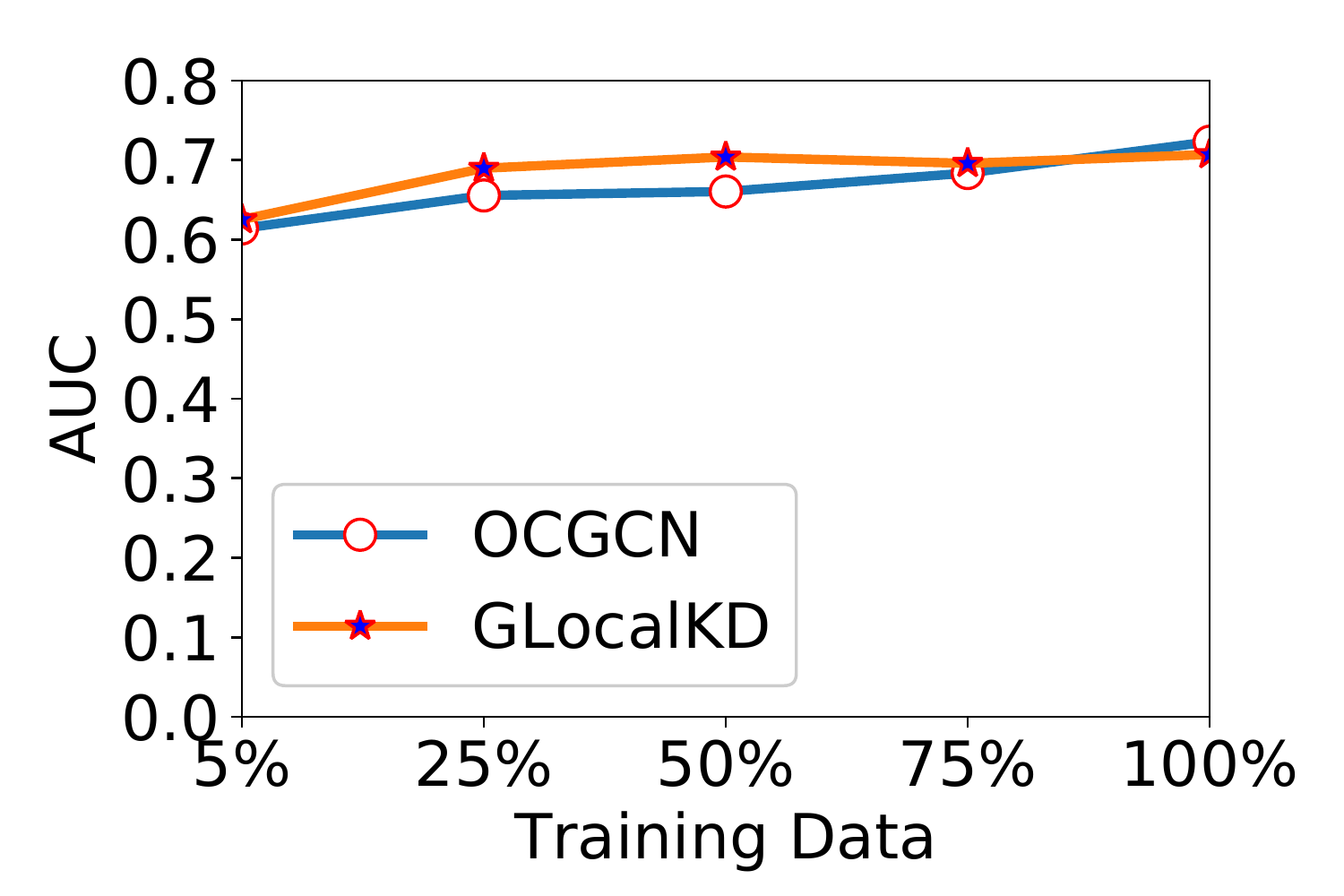}}\hspace{-6pt}
  \subfigure[COX2]{\includegraphics[width=4.25cm]{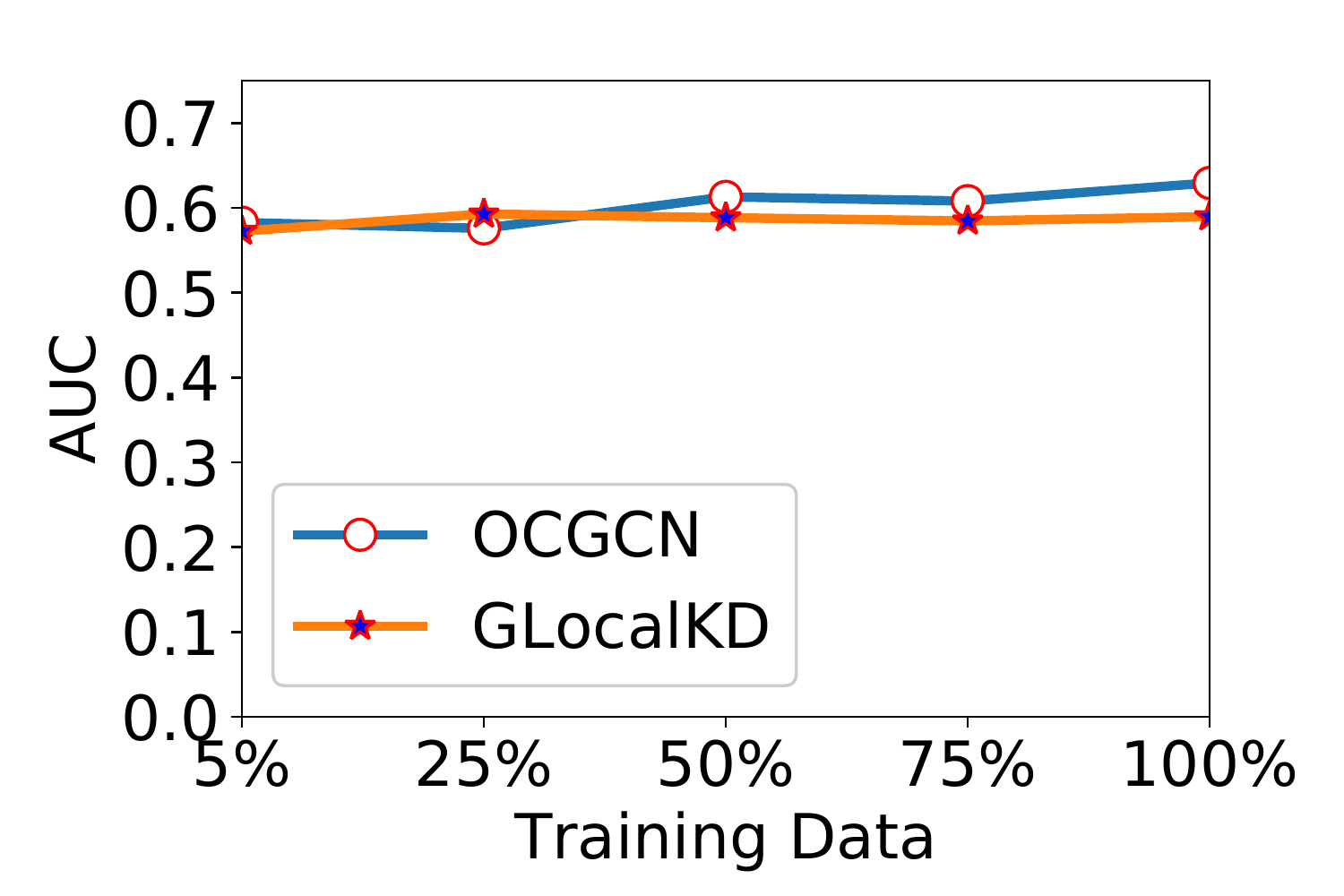}}
  \caption{AUC performance of GLocalKD and OCGCN using different amount of training data.}
  \label{dataefficiency}
\end{figure}

The AUC results are shown in Figure~\ref{dataefficiency}. It is very impressive that even when 95\% less training data are used, GLocalKD can retain similarly good performance across nearly all the six datasets. By contrast, the performance of OCGCN can drop significantly on some datasets, such as ENZYMES and AIDS, if the same amount of training data is reduced. As a result, GLocalKD can outperform OCGCN by large margins even it uses 95\% less training data than OCGCN on such datasets.

\subsection{Robustness w.r.t. Anomaly Contamination}\label{subsec:robustness}
\subsubsection{Experiment Settings} 
Recall that we tackle the semi-supervised anomaly detection setting with exclusively normal training samples. However, the data collected in real applications may be contaminated by some anomalies or data noises. 
This section investigates the robustness of GLocalKD w.r.t. different anomaly contamination levels in the training data. We vary the contamination rates from $0\%$ up to $16\%$. Again, we report the results on the six attributed graph datasets only due to page limitation; OCGCN is used as baseline.

\subsubsection{Findings}
\begin{figure}[t!]
  \centering
  \subfigcapskip=-4pt
  \subfigbottomskip=-6pt
  \subfigure[PROTEINS$\_$full]{\includegraphics[width=4.25cm]{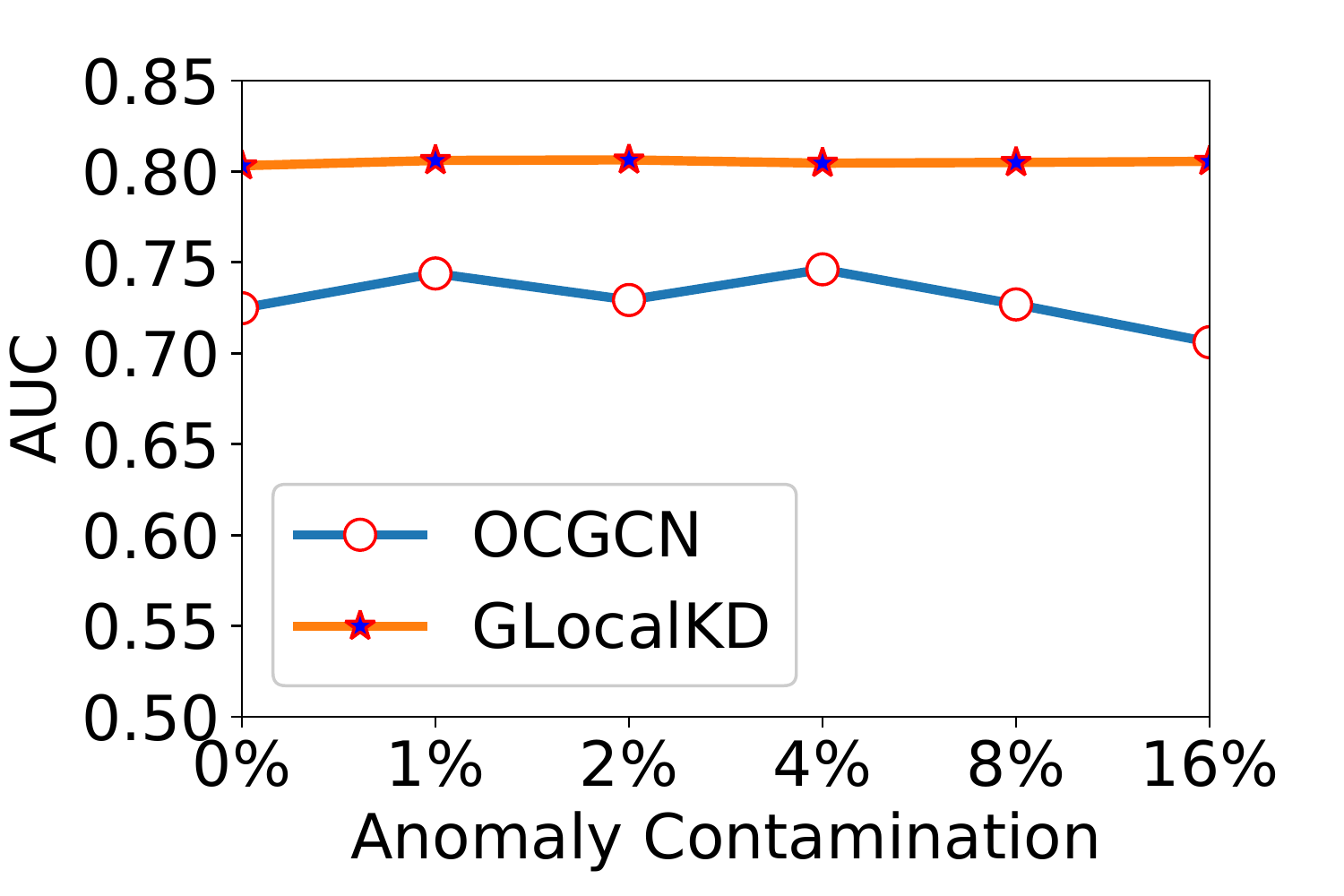}}\hspace{-6pt}
  \subfigure[ENZYMES]{\includegraphics[width=4.25cm]{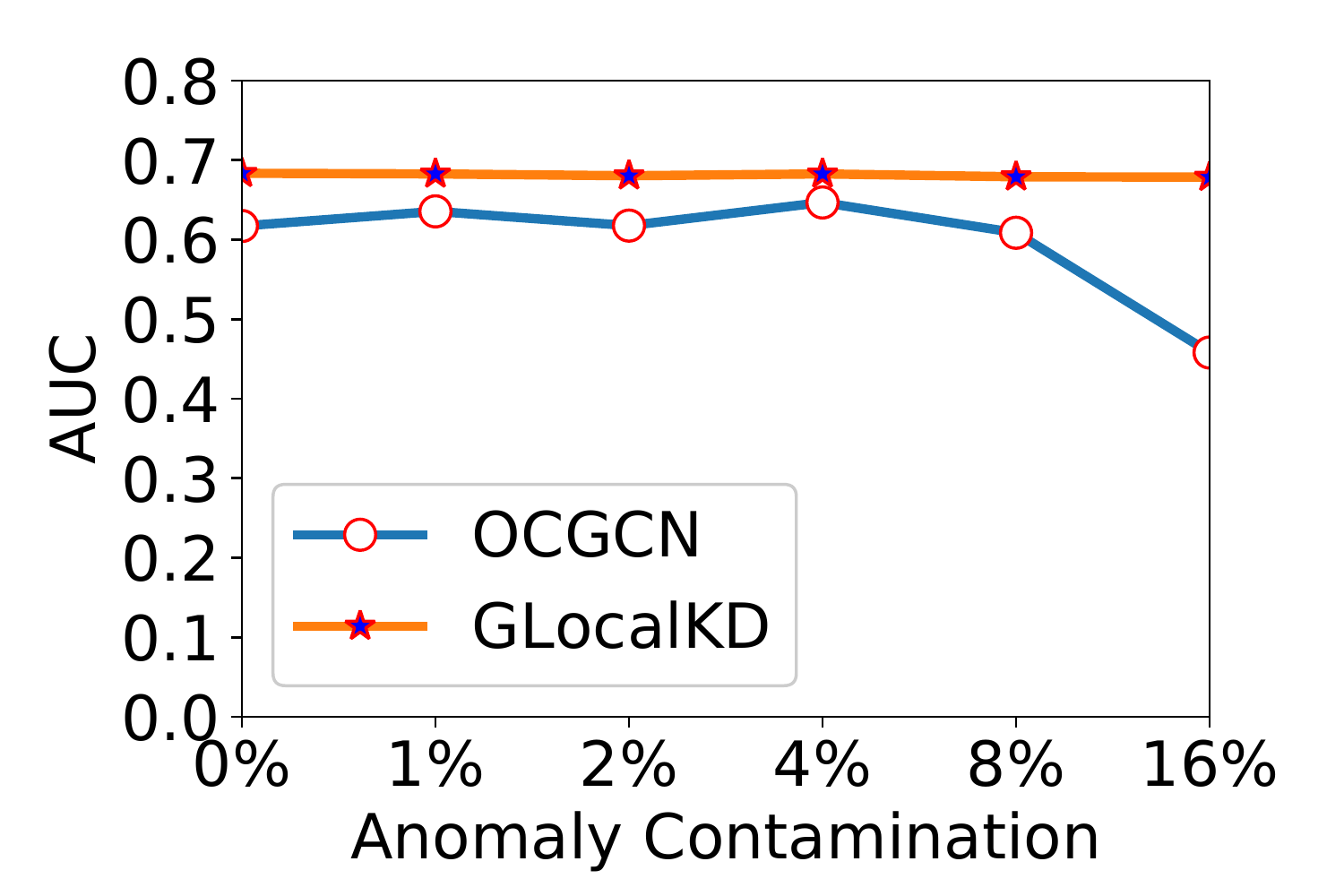}}
  \subfigure[AIDS]{\includegraphics[width=4.25cm]{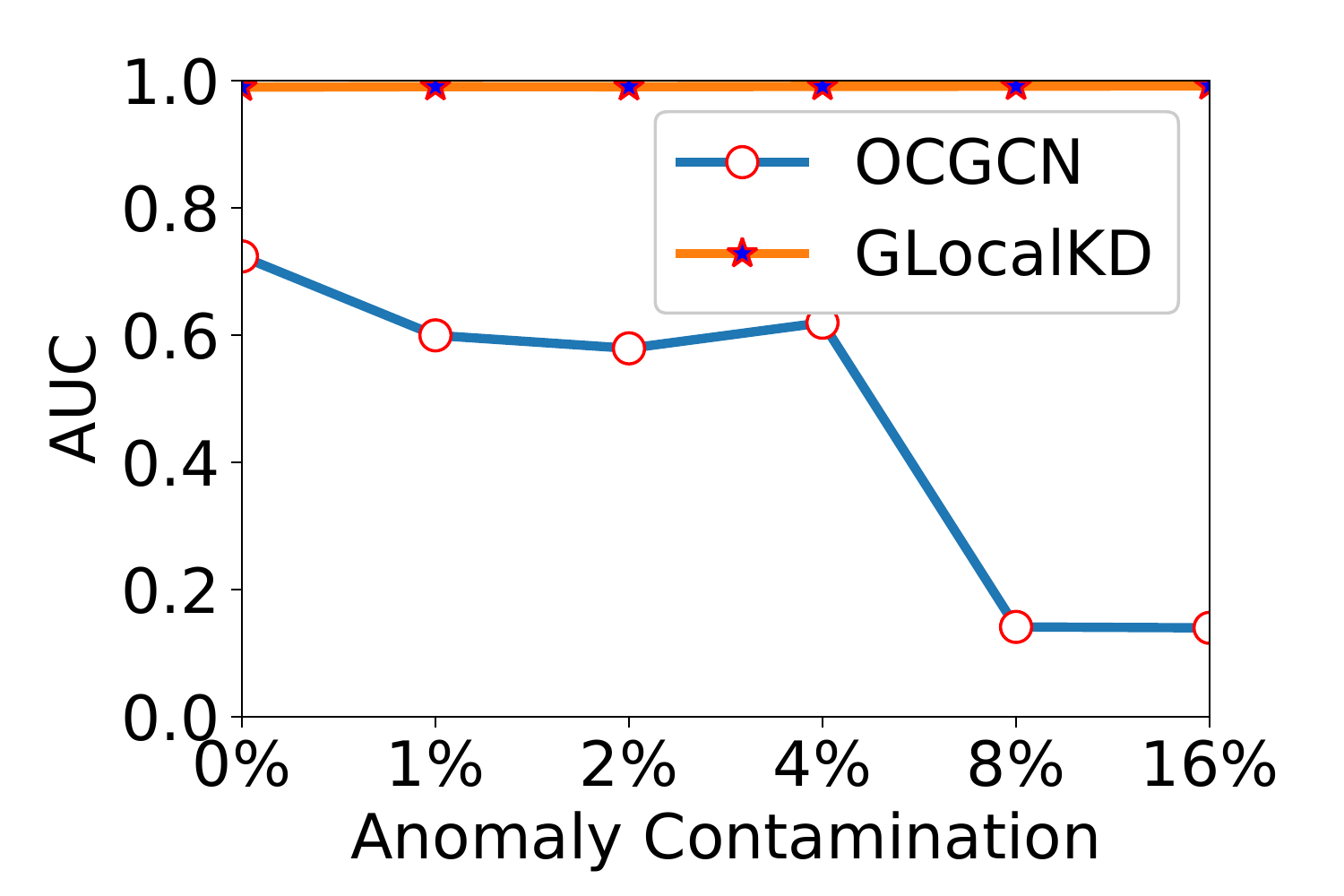}}\hspace{-6pt}
  \subfigure[DHFR]{\includegraphics[width=4.25cm]{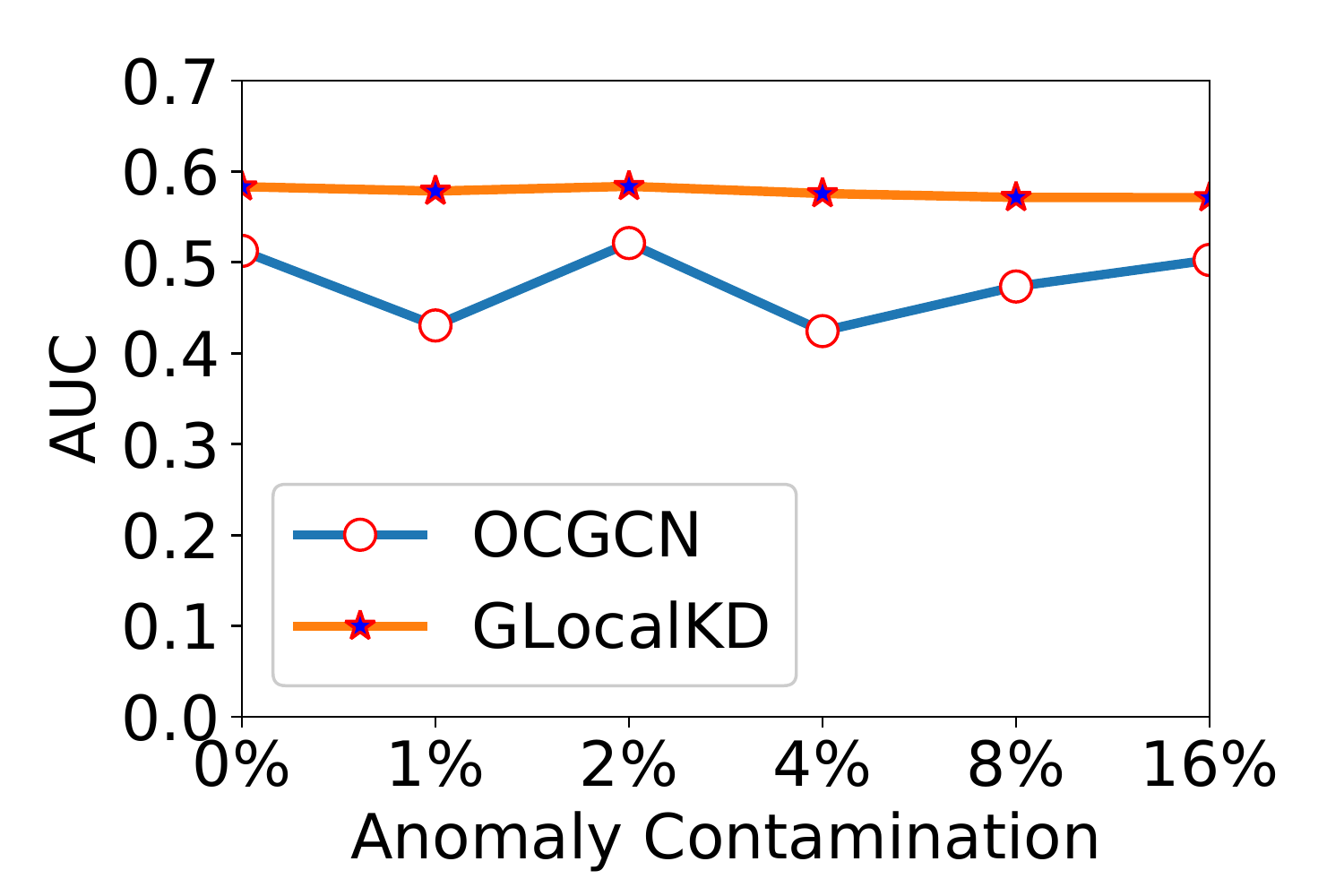}}
  \subfigure[BZR]{\includegraphics[width=4.25cm]{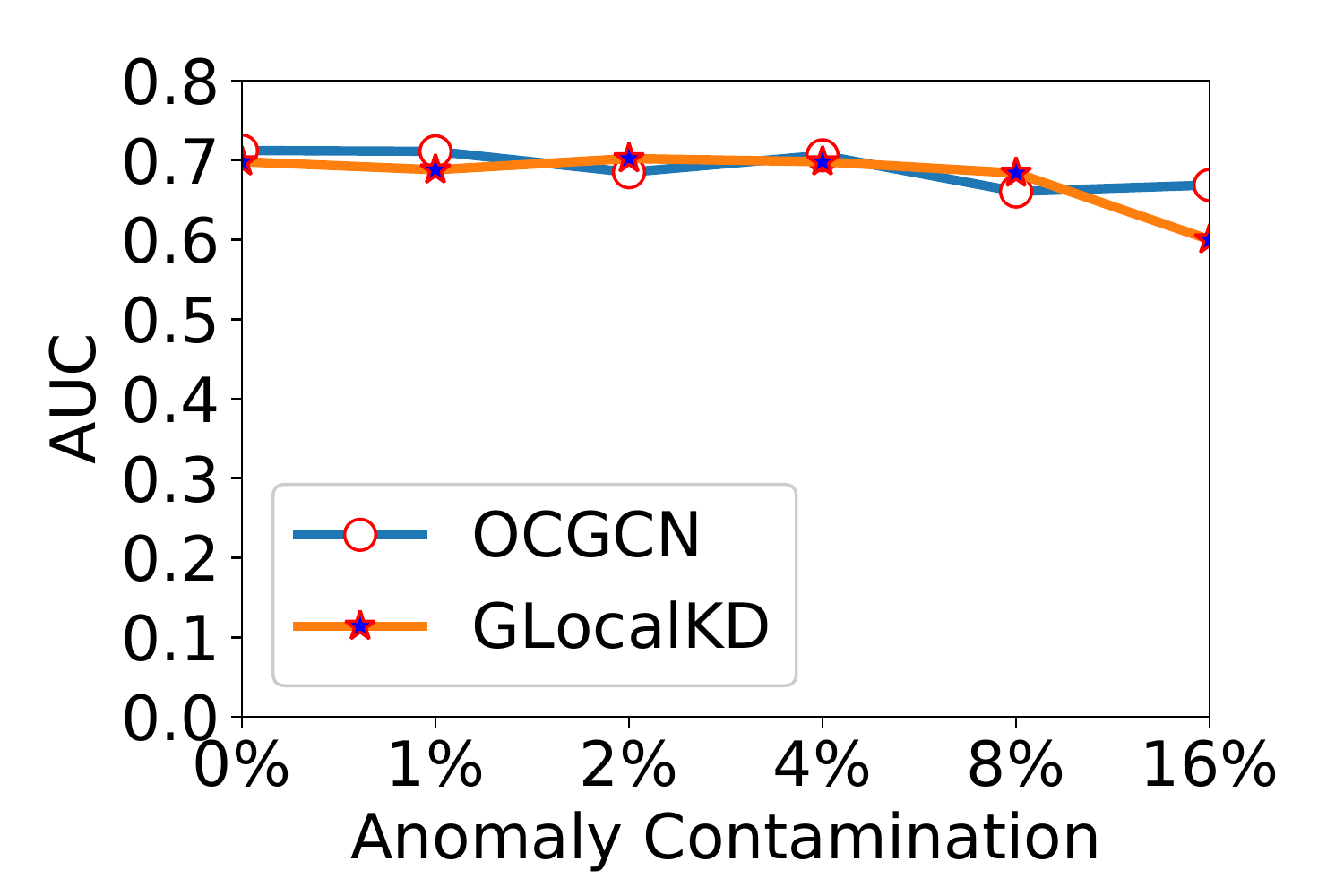}}\hspace{-6pt}
  \subfigure[COX2]{\includegraphics[width=4.25cm]{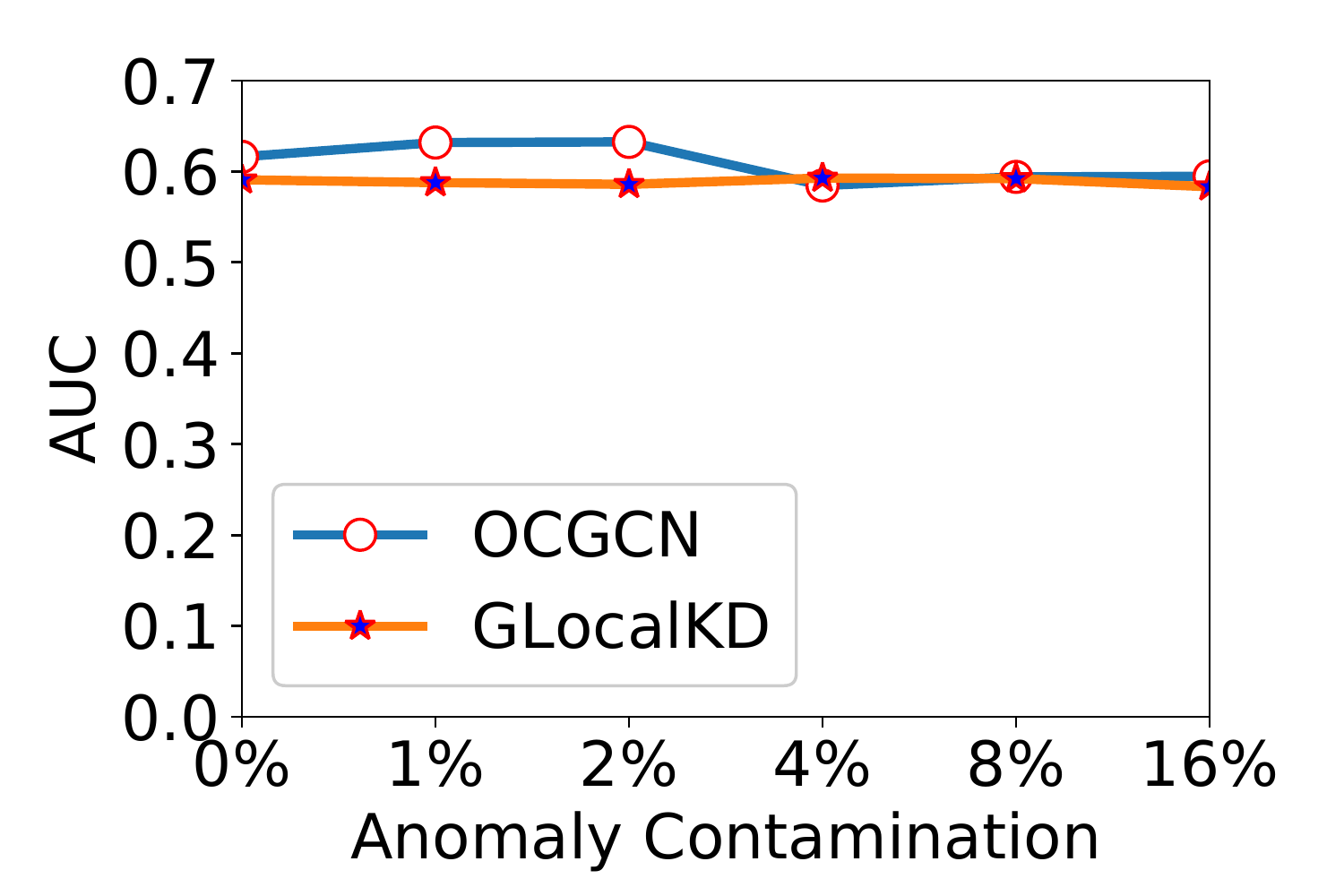}}
  \caption{AUC performance of GLocalKD and OCGCN w.r.t. different anomaly contamination rates.}
  \label{robustness}
\end{figure}

AUC results of GLocalKD and OCGCN with different anomaly contamination rates are shown in Figure~\ref{robustness}. GLocalKD is barely affected by the contamination and performs very stably on all the datasets, contrasting to OCGCN whose performance decreases largely with increasing contamination rate on ENZYMES and AIDS. This is mainly because GLocalKD essentially learns all types of patterns in the training data by the random distillation, by which it is able to detect the anomalies as long as those anomalous patterns are not as frequent as the normal patterns in the training data; whereas OCGCN is sensitive since its anomaly measure, SVDD, is sensitive to the anomaly contamination.

\subsection{Sensitivity Test}\label{subsec:sensitivity}
\subsubsection{Experiment Settings}
This section tests the sensitivity of GLocalKD to the representation dimension and the GCN depth. For the first test, we vary the output dimension of GCN in $\{32, 64, 128, 256, 512\}$; for the GCN depth, we evaluate the performance of GLocalKD using $k$ GCN layers, with $k\in \{1,2,3,5\}$. The results are illustrated in Figures \ref{dimen} and \ref{layer} in Appendix \ref{subsec:sensitivitytestAPP}.

\subsubsection{Sensitivity}
As can be seen from the results, GLocalKD performs stably using different representation dimensionality sizes on most datasets. 
The dimensionality size -- 256 -- is generally recommended as this setting enables GLocalKD to perform well on diverse datasets.

Besides, GLocalKD achieves better performance with increasing depth on nearly all the datasets, but the performance is flatten when increasing the depth from three to five. A network depth of three is generally recommended, since deeper GCN does not help achieve better performance but is more computationally costly.

\subsection{Ablation Study}\label{subsec:ablation}
\subsubsection{Experiment Settings} \label{subsec:loss terms}
In this section, we examine the importance of the two components, $L_{\mathit{graph}}$ and $L_{\mathit{node}}$, in our model. 
To do that, we derive two variants of GLocalKD, including GLocalKD w/o $L_{\mathit{node}}$ that denotes the use of random distillation on the graph representations only, and GLocalKD w/o $L_{\mathit{graph}}$ that represents the use of random distillation on the node representations only.
\subsubsection{Findings}
The results of GLocalKD and its two variants are shown in Table~\ref{term}. It is clear that using $L_{\mathit{graph}}$ (or $L_{\mathit{node}}$) only can obtain better performance on some datasets, while it may perform worse on the other datasets, compared with GLocalKD. Joint random distillation by using both of $L_{\mathit{graph}}$ and $L_{\mathit{node}}$ can achieve a good trade-off and perform generally good across all the datasets. 

It is interesting that GLocalKD w/o $L_{\mathit{graph}}$ significantly outperforms GLocalKD w/o $L_{\mathit{node}}$ in a number of datasets, \eg, AIDS, DHFR, DD, MMP, p53, PPAR-gamma and hERG, indicating the dominant presence of locally-anomalous graphs in those data; on the other hand, the inverse cases occur on ENZYMES, IMDB and HSE, indicating the dominance of globally-anomalous graphs in these three datasets. These results show that modeling fine-grained graph regularity is as important as, if not more important than, the holistic graph regularity for the GAD task, since both types of graph anomalies can present in the graph datasets.

\begin{table}[h]
\caption{Detection of locally/globally-anomalous graphs.}
\label{term}
\scalebox{0.9}{
\begin{tabular}{lccc}
\toprule  
\textbf{Dataset} & \textbf{GLocalKD} & \textbf{w/o $L_{\mathit{node}}$} & \textbf{w/o $L_{\mathit{graph}}$}\\
\midrule
PROTEINS$\_$full & \textbf{0.785}$\pm$0.034 & 0.686$\pm$0.045 & 0.757$\pm$0.040\\
ENZYMES & 0.636$\pm$0.061 & \textbf{0.642}$\pm$0.096 & 0.505$\pm$0.036\\
AIDS & 0.992$\pm$0.004 & 0.963$\pm$0.014 & \textbf{0.997}$\pm$0.006\\
DHFR & 0.558$\pm$0.030 & 0.459$\pm$0.036 & \textbf{0.596}$\pm$0.030\\
BZR & \textbf{0.679}$\pm$0.065 & 0.623$\pm$0.079 & 0.671$\pm$0.049\\
COX2 & \textbf{0.589}$\pm$0.045 & 0.585$\pm$0.051 & 0.557$\pm$0.055\\
DD & 0.805$\pm$0.017 & 0.528$\pm$0.093 & \textbf{0.805}$\pm$0.017\\
NCI1 & \textbf{0.683}$\pm$0.015 & 0.458$\pm$0.058 & 0.682$\pm$0.015\\
IMDB & 0.514$\pm$0.039 & \textbf{0.610}$\pm$0.103 & 0.490$\pm$0.044\\
REDDIT & \textbf{0.782}$\pm$0.016 & 0.574$\pm$0.085 & 0.781$\pm$0.016\\
HSE & 0.591$\pm$0.001 & \textbf{0.655}$\pm$0.007 & 0.589$\pm$0.000\\
MMP & 0.676$\pm$0.001 & 0.543$\pm$0.016 & \textbf{0.680}$\pm$0.000\\
p53 & 0.639$\pm$0.002 & 0.495$\pm$0.016 & \textbf{0.641}$\pm$0.000\\
PPAR-gamma & 0.644$\pm$0.001 & 0.600$\pm$0.044 & \textbf{0.644}$\pm$0.000\\
COLLAB & 0.525$\pm$0.014 & 0.501$\pm$0.055 & \textbf{0.526}$\pm$0.012\\
hERG  & \textbf{0.704}$\pm$0.049 & 0.566$\pm$0.043 & 0.703$\pm$0.057\\
\bottomrule
\end{tabular}
}
\end{table}

\section{Conclusion}
This paper proposes a novel framework and its instantiation GLocalKD to detect abnormal graphs in a set of graphs. As shown in our experimental results, graph datasets can contain different types of anomalies -- locally- and globally-anomalous graphs. To the best of our knowledge, GLocalKD is the first model designed to detect both types of graph anomalies. Extensive experiments demonstrate that GLocalKD performs significantly better in AUC and can be trained much more sample-efficiently when compared with its advanced counterparts. We also show that GLocalKD achieves promising AUC performance even when there is large anomaly contamination in the training data, indicating that GLocalKD can be applied in not only semi-supervised settings (exclusively normal training data) but also unsupervised settings (anomaly-contaminated unlabeled training data).  

\begin{acks}
In this work R. Ma and L. Chen are supported by ARC DP210101347.
\end{acks}

\newpage

\bibliographystyle{ACM-Reference-Format}
\balance
\bibliography{reference}

\newpage

\section*{Appendix}
\appendix

In the appendix, we provide more detailed information about the implementation details of our model GLocalKD and its competing methods, as well as some extra empirical results. The algorithmic procedure of GLocalKD is presented in Appendix \ref{subsec:pseudocode}, while the GNN architecture, hyperparameter settings and optimization are presented in Appendix \ref{subsec:implementation}. Appendix \ref{subsectime} presents the training and test time of GLocalKD and its competitors on three representative datasets. Appendix \ref{subsec:sensitivitytestAPP} shows the sensitivity test results. Lastly, Appendix ~\ref{LESINN} presents the influence of subsample size on the performance of LESINN. 

\section{The Algorithm of GLocalKD}\label{subsec:pseudocode}
Algorithm~\ref{glocalkd} presents the procedure of training GLocalKD. After random weight initialization of $\hat{\mathbf{\Theta}}$ and $\mathbf{\Theta}$ in Step 1, GLocalKD performs stochastic gradient descent-based optimization to learn $\mathbf{\Theta}$ of the predictor network in Steps 2-11, while the parameters in $\hat{\mathbf{\Theta}}$ are fixed. Particularly, Step 4 samples a mini-batch $\mathcal{B}$ with size $batch\_size$. We obtain node representations and graph representations from both of $\hat{\phi}(\cdot, \hat{\mathbf{\Theta}})$ and $\phi(\cdot,\mathbf{\Theta})$ in Steps 6-7, respectively.
Step 9 then performs gradient descent steps on our loss Eq. 12 w.r.t. the parameters in $\mathbf{\Theta}$. We finally obtain the predictor network $\phi(\cdot,\mathbf{\Theta}^*)$ with the learned $\mathbf{\Theta}^*$ and the random target network $\hat{\phi}(\cdot,\hat{\mathbf{\Theta}})$.

\renewcommand{\algorithmicrequire}{\textbf{Input:}}
\renewcommand{\algorithmicensure}{\textbf{Output:}}
\begin{algorithm}[h]
    \caption{Training GLocalKD}\label{glocalkd} 
    \begin{algorithmic}[1]
    \REQUIRE Normal training graph set $\mathcal{G}=\{G_i\}_i$ 
    \ENSURE Target network -- $\hat{\phi}(\cdot,\hat{\mathbf{\Theta}})$, predictor network -- $\phi(\cdot,\mathbf{\Theta}^*)$
        \STATE Randomly initialize $\hat{\mathbf{\Theta}}$ and $\mathbf{\Theta}$, with $\hat{\mathbf{\Theta}}$ fixed
        \FOR{$i=1$ to $n\_epochs$}
            \FOR{$j=1$ to $n\_batches$}
            \STATE $\mathcal{B}\leftarrow$Randomly sample $batch\_size$ graphs from $\mathcal{G}$ 
            \FOR{$G$ in $\mathcal{B}$} 
                \STATE Compute node representations $\hat{\mathbf{h}}_i$ and $\mathbf{h}_i$, $\forall v_i \in \mathcal{V}_G$
                \STATE Compute graph representations $\hat{\mathbf{h}}_G$ and $\mathbf{h}_G$
            \ENDFOR
            \STATE Perform a gradient descent step on Eq. 12 w.r.t. the parameters in $\mathbf{\Theta}$
            \ENDFOR
        \ENDFOR
        \RETURN $\hat{\phi}(\cdot,\hat{\mathbf{\Theta}})$, $\phi(\cdot,\mathbf{\Theta}^*)$
    \end{algorithmic}
\end{algorithm}

\section{Implementation Details}\label{subsec:implementation}
All experiments are carried out on NVIDIA Quadro RTX 6000 GPU with Intel Xeon E-2288G 3.7GHz CPU, and all models are implemented with Python 3.6\footnote{https://www.python.org/}.

The target network and the predictor network in GLocalKD share the same network architecture -- a network with three GCN layers. The dimension of the hidden layer is 512 and the output layer has 256 neural units. 
As indicated in Eq. 9, max pooling is used to obtain the graph representations.
The GCN weight parameters are initialized using the Kaiming uniform method, with the bias parameters initialized to be zeros. For attributed graph datasets, the feature matrix $\mathbf{X}$ is directly built upon their node features; for datasets with plain graphs, the degree of each node is used as the node features. 
The learning rate is set to $10^{-4}$ by default except on the ENZYMES, AIDS, DHFR, HSE, p53, MMP and PPAR-gamma datasets where the learning rate is set to $10^{-5}$. Nevertheless, the performance of GLocalKD has small variations on these datasets either using $10^{-4}$ or $10^{-5}$ as the learning rate. On the dataset PROTEINS$\_$full, the learning rate is set to $10^{-2}$ to obtain good performance. 
The batch size is 300 for all data sets except the four largest datasets HSE, MMP, p53 and PPAR-gamma, for which the batch size is 2000. The number of epochs is 150 for all data sets.

\begin{table}[htbp]
\caption{Training and test time on three datasets: REDDIT, p53 and COLLAB.}
\label{timetable}
\setlength{\tabcolsep}{0.6mm}
\scalebox{0.65}{
\begin{tabular}{l|c|cccccccc}
\hline
& \multirow{2}*{\textbf{Dataset}} &  \multicolumn{2}{c}{\textbf{InfoGraph}} & \multicolumn{2}{c}{\textbf{WL}} & \multicolumn{2}{c}{\textbf{PK}} & \multirow{2}*{\textbf{OCGCN}} & \multirow{2}*{\textbf{GLocalKD}} \\ 
\cline{3-8}
 &  & \textbf{iForest} & \textbf{LESINN} & \textbf{iForest} & \textbf{LESINN} &  \textbf{iForest} & \textbf{LESINN} & &  \\
\hline
\multirow{3}*{Training Time} & REDDIT &  3536.36 & 3536.36 & 8.01 & 8.01 & 127.09 & 127.09 & 1397.40 & 1395\\
& p53 & 60.61 & 60.61 & 9.42 & 9.42 & 821.29 & 821.29 & 297.37& 337.80 \\
& COLLAB & 2059.66 & 2059.66 &  63.24 & 63.24 & 416.95& 416.95 & 2421.83& 2510.52\\
\hline
\multirow{3}*{Test Time} & REDDIT & 5.79 & 15.19 & 3.72 & 29.37 & 84.46 & 112.20 & 4.65 & 4.97 \\
& p53 & 19.67 & 24.54 & 225.44 & 207.95 & 301.89 & 250.79 & 0.66 & 0.97 \\
& COLLAB & 12.41& 34.44 & 39.50 & 313.65 & 273.82 & 573.90 & 9.28 & 8.88\\
\hline
\end{tabular}
}
\end{table}

The architecture of GCN used in OCGCN is exactly the same as our model. The learning rate is also searched from $10^{-1}$ to $10^{-5}$. 
We use the same method in Deep SVDD [32] to generate the one-class center. Both of GLocalKD and OCGCN are implemented using Pytorch 1.9\footnote{https://pytorch.org/}.
Similarly, InfoGraph is taken from its official implementation\footnote{https://github.com/fanyun-sun/InfoGraph}, which uses a three-layer GIN architecture, with the learning rate set to $10^{-3}$. Adam is the default optimizer used in the above three methods.
Both WL and PK are directly taken from the GraKeL library 0.1.8\footnote{https://github.com/ysig/GraKeL\label{grakel}}. For WL, we perform three iterations to obtain the graph representations, which utilize the same neighborhood information as a three-layer GCN as in our model and OCGCN. PK is used with the recommended setting in GraKeL. For iForest\footnote{https://github.com/scikit-learn/} in each method, we adjust its parameters, including the number of trees, subsample size and contamination rate. We finally use the recommended settings as in \cite{liu2008isolation}, \ie 100 for the number of trees, 256 for subsampling size  and 0.1 for contamination rate, since the results of iForest do not change much with varying hyperparameter settings. Two parameters in LESINN\footnote{https://github.com/GuansongPang/deep-outlier-detection/}, \ie, ensemble size and subsample size, are chosen from $\{2,4,8,16,32,64,128,256\}$. In Table \ref{auroc}, we report the results of LESINN by setting both of these two parameters to 256, as this setting enables the most effective results across all datasets (see Table \ref{lesinn}).

\begin{table*}[t!]
\caption{Results of LESINN using different subsampling sizes.}
\label{lesinn}
\setlength{\tabcolsep}{0.6mm}
\scalebox{0.75}{
\begin{tabular}{l|c|cccccccc}
\hline
Dataset&	Method&	subsample size=2 &	subsample size=4 &	subsample size=8&	subsample size=16&	subsample size=32&	subsample size=64&	subsample size=128&	subsample size=256 \\
\hline
\multirow{3}*{PROTEINS$\_$full} &	Info-LESINN &	0.399$\pm$0.048&	0.398$\pm$0.049&	0.390$\pm$0.049&	0.380$\pm$0.049&	0.367$\pm$0.047&	0.357$\pm$0.047&	0.345$\pm$0.048&	0.336$\pm$0.047 \\
&	WL-LESINN &	0.769$\pm$0.017&	0.779$\pm$0.014&	0.780$\pm$0.014&	0.779$\pm$0.016&	0.777$\pm$0.021&	0.769$\pm$0.029&	0.742$\pm$0.047&	0.712$\pm$0.053 \\
&	PK-LESINN &	 0.759$\pm$0.023&	0.766$\pm$0.024&	0.769$\pm$0.020&	0.769$\pm$0.018&	0.765$\pm$0.019&	0.702$\pm$0.068&	 0.633$\pm$0.057&	0.572$\pm$0.031 \\
\multirow{3}*{ENZYMES} &	Info-LESINN &	0.465$\pm$0.078&	0.465$\pm$0.072&	0.462$\pm$0.059&	0.465$\pm$0.051&	0.466$\pm$0.042&	0.477$\pm$0.041&	0.496$\pm$0.044&	0.528$\pm$0.046 \\
&	WL-LESINN &	0.519$\pm$0.066&	0.488$\pm$0.069&	0.485$\pm$0.065&	0.498$\pm$0.050&	0.518$\pm$0.040&	0.538$\pm$0.037&	0.577$\pm$0.044&	0.624$\pm$0.050 \\
&	PK-LESINN &	0.562$\pm$0.053&	0.553$\pm$0.041&	0.564$\pm$0.030&	0.579$\pm$0.025&	0.590$\pm$0.034&	0.596$\pm$0.043&	 0.594$\pm$0.038&	0.608$\pm$0.033 \\
\multirow{3}*{AIDS} &	Info-LESINN &	0.883$\pm$0.041&	0.889$\pm$0.039&	0.900$\pm$0.038&	0.912$\pm$0.035&	0.924$\pm$0.033&	0.935$\pm$0.030&	0.944$\pm$0.027&	0.955$\pm$0.023 \\
&	WL-LESINN &	0.651$\pm$0.016&	0.526$\pm$0.021&	 0.437$\pm$0.019&	0.424$\pm$0.012&	0.447$\pm$0.016&	0.483$\pm$0.013&	 0.528$\pm$0.014&	0.584$\pm$0.016 \\
&	PK-LESINN &	0.578$\pm$0.026&	0.468$\pm$0.037&	0.385$\pm$0.021&	0.352$\pm$0.017&	0.358$\pm$0.010&	0.374$\pm$0.009&	0.393$\pm$0.010&	0.421$\pm$0.010 \\
\multirow{3}*{DHFR}&	Info-LESINN &	0.460$\pm$0.042&	0.473$\pm$0.046&	0.486$\pm$0.040&	0.509$\pm$0.042&	0.541$\pm$0.039&	0.575$\pm$0.035&	0.608$\pm$0.034&	0.625$\pm$0.028 \\
&	WL-LESINN &	0.365$\pm$0.038&	0.401$\pm$0.047&	 0.457$\pm$0.049&	0.480$\pm$0.048&	0.509$\pm$0.052&	0.538$\pm$0.055&	0.573$\pm$0.059&	0.596$\pm$0.056 \\
&	PK-LESINN &	 0.368$\pm$0.032&	0.400$\pm$0.027&	0.431$\pm$0.021&	0.453$\pm$0.027&	0.474$\pm$0.040&	0.503$\pm$0.051&	0.541$\pm$0.056&	0.568$\pm$0.054 \\
\multirow{3}*{BZR} &	Info-LESINN &	0.557$\pm$0.043&	0.568$\pm$0.037&	0.600$\pm$0.039&	0.632$\pm$0.039&	0.658$\pm$0.040&	0.690$\pm$0.050&	0.721$\pm$0.068&	0.737$\pm$0.071 \\
&	WL-LESINN &	0.540$\pm$0.054&	0.549$\pm$0.050&	0.576$\pm$0.061&	0.620$\pm$0.055&	0.679$\pm$0.053&	0.700$\pm$0.050&	0.717$\pm$0.043&	 0.720$\pm$0.032 \\
&	PK-LESINN &	 0.528$\pm$0.070&	0.542$\pm$0.067&	0.578$\pm$0.075&	0.631$\pm$0.073&	0.693$\pm$0.072&	0.739$\pm$0.068&	0.764$\pm$0.066&	0.775$\pm$0.063 \\
\multirow{3}*{COX2} &	Info-LESINN &	0.588$\pm$0.064&	0.611$\pm$0.050&	0.616$\pm$0.052&	0.628$\pm$0.058&	0.639$\pm$0.066&	0.661$\pm$0.069&	0.673$\pm$0.066&	0.670$\pm$0.079 \\
&	WL-LESINN &	0.444$\pm$0.101&	0.487$\pm$0.089&	0.512$\pm$0.074&	 0.557$\pm$0.075&	0.583$\pm$0.073&	0.599$\pm$0.074&	 0.605$\pm$0.067&	0.590$\pm$0.056 \\
&	PK-LESINN &	0.443$\pm$0.093&	0.465$\pm$0.085&	 0.472$\pm$0.080&	0.523$\pm$0.073&	 0.568$\pm$0.067&	0.608$\pm$0.061&	0.648$\pm$0.046&	0.671$\pm$0.039 \\
\multirow{3}*{DD} &	Info-LESINN &	0.320$\pm$0.038&	0.318$\pm$0.033&	0.315$\pm$0.032&	0.313$\pm$0.031&	0.310$\pm$0.032&	0.308$\pm$0.032&	0.307$\pm$0.032&	0.310$\pm$0.034 \\
&	WL-LESINN &	0.543$\pm$0.052&	0.540$\pm$0.048&	0.535$\pm$0.050&	 0.547$\pm$0.055&	 0.560$\pm$0.054&	0.578$\pm$0.055&	0.605$\pm$0.051&	0.638$\pm$0.045 \\
&	PK-LESINN &	 0.800$\pm$0.023&	0.811$\pm$0.028&	0.816$\pm$0.028&	0.819$\pm$0.027&	0.822$\pm$0.026&	0.827$\pm$0.027&	0.831$\pm$0.025&	0.833$\pm$0.023 \\
\multirow{3}*{NCI1} &	Info-LESINN &	0.479$\pm$0.016&	0.482$\pm$0.018&	0.487$\pm$0.019&	0.495$\pm$0.023&	0.508$\pm$0.027&	0.532$\pm$0.031&	0.561$\pm$0.034&	0.598$\pm$0.035 \\
&	WL-LESINN &	0.533$\pm$0.029&	0.566$\pm$0.029&	0.590$\pm$0.024&	0.621$\pm$0.019&	0.650$\pm$0.015&	0.676$\pm$0.014&	0.710$\pm$0.014&	 0.743$\pm$0.015 \\
&	PK-LESINN &	0.525$\pm$0.021&	0.542$\pm$0.024&	0.558$\pm$0.019&	0.586$\pm$0.019&	 0.607$\pm$0.017&	0.624$\pm$0.015&	0.646$\pm$0.013&	0.670$\pm$0.012 \\
\multirow{3}*{IMDB} &	Info-LESINN &	0.431$\pm$0.033&	0.438$\pm$0.033&	0.441$\pm$0.029&	0.467$\pm$0.045&	0.482$\pm$0.043&	0.505$\pm$0.037&	0.541$\pm$0.023&	0.565$\pm$0.017 \\
&	WL-LESINN &	0.398$\pm$0.040&	0.397$\pm$0.028&	0.404$\pm$0.028&	0.437$\pm$0.027&	0.504$\pm$0.055&	0.586$\pm$0.058&	0.605$\pm$0.057&	0.612$\pm$0.046 \\
&	PK-LESINN &	0.392$\pm$0.045&	0.384$\pm$0.037&	0.385$\pm$0.033&	0.406$\pm$0.023&	0.462$\pm$0.045&	0.552$\pm$0.057&	0.582$\pm$0.050&	0.585$\pm$0.047 \\
\multirow{3}*{REDDIT} &	Info-LESINN &	0.449$\pm$0.023&	0.461$\pm$0.030&	0.418$\pm$0.038&	0.346$\pm$0.048&	0.290$\pm$0.032&	0.276$\pm$0.028&	0.268$\pm$0.027&	0.262$\pm$0.027 \\
&	WL-LESINN &	0.231$\pm$0.026&	0.234$\pm$0.026&	0.237$\pm$0.027&	 0.239$\pm$0.027&	0.239$\pm$0.027&	 0.239$\pm$0.028&	0.239$\pm$0.028&	0.239$\pm$0.028 \\
&	PK-LESINN &	0.224$\pm$0.024&	0.295$\pm$0.035&	0.422$\pm$0.017&	0.440$\pm$0.013&	0.448$\pm$0.013&	0.457$\pm$0.010&	0.471$\pm$0.010&	0.487$\pm$0.013 \\
\multirow{3}*{HSE} &	Info-LESINN &	0.586$\pm$0.116&	0.589$\pm$0.107&	0.596$\pm$0.100&	0.606$\pm$0.092&	0.617$\pm$0.083&	0.629$\pm$0.071&	0.644$\pm$0.060&	0.657$\pm$0.051 \\
&	WL-LESINN &	0.341$\pm$0.000&	0.421$\pm$0.000&	0.468$\pm$0.000&	0.482$\pm$0.000&	0.495$\pm$0.000&	0.507$\pm$0.000&	0.518$\pm$0.000&	0.528$\pm$0.000 \\
&	PK-LESINN &	0.361$\pm$0.005&	0.393$\pm$0.011&	0.407$\pm$0.011&	0.419$\pm$0.006&	0.435$\pm$0.004&	 0.446$\pm$0.008&	 0.462$\pm$0.013&	0.469$\pm$0.016 \\
\multirow{3}*{MMP} &	Info-LESINN &	0.626$\pm$0.051&	 0.612$\pm$0.051&	0.600$\pm$0.048&	0.587$\pm$0.043&	0.579$\pm$0.039&	0.574$\pm$0.038&	0.571$\pm$0.038&	0.571$\pm$0.037 \\
&	WL-LESINN &	0.422$\pm$0.000&	0.363$\pm$0.000&	0.344$\pm$0.000&	0.333$\pm$0.000&	0.330$\pm$0.000&	0.320$\pm$0.000&	0.313$\pm$0.000&	0.307$\pm$0.000 \\
&	PK-LESINN &	0.400$\pm$0.010&	0.362$\pm$0.002&	0.354$\pm$0.004&	0.348$\pm$0.004&	0.341$\pm$0.006&	0.332$\pm$0.005&	0.326$\pm$0.007&	 0.322$\pm$0.008 \\
\multirow{3}*{p53} &	Info-LESINN &	0.573$\pm$0.046&	0.567$\pm$0.045&	0.551$\pm$0.041&	0.537$\pm$0.037&	 0.532$\pm$0.033&	0.525$\pm$0.030&	0.520$\pm$0.028&	0.520$\pm$0.025 \\
&	WL-LESINN &	0.435$\pm$0.000&	0.429$\pm$0.000&	0.413$\pm$0.000&	0.413$\pm$0.000&	0.409$\pm$0.000&	0.403$\pm$0.000&	0.396$\pm$0.000&	0.390$\pm$0.000 \\
&	PK-LESINN &	0.341$\pm$0.007&	0.342$\pm$0.005&	0.340$\pm$0.004&	0.339$\pm$0.003&	0.339$\pm$0.004&	0.336$\pm$0.003&	0.332$\pm$0.001&	0.329$\pm$0.001 \\
\multirow{3}*{PPAR-gamma} &	Info-LESINN &	0.629$\pm$0.026&	0.625$\pm$0.038&	0.616$\pm$0.042&	0.605$\pm$0.044&	0.594$\pm$0.049&	0.574$\pm$0.049&	0.553$\pm$0.043&	0.541$\pm$0.036 \\
&	WL-LESINN &	0.379$\pm$0.000&	0.409$\pm$0.000&	0.428$\pm$0.000&	0.444$\pm$0.000&	0.460$\pm$0.000&	0.461$\pm$0.000&	0.458$\pm$0.000&	0.461$\pm$0.000 \\
&	PK-LESINN &	0.408$\pm$0.005&	0.405$\pm$0.006&	0.404$\pm$0.007&	0.400$\pm$0.012&	0.400$\pm$0.014&	0.397$\pm$0.016&	0.388$\pm$0.015&	0.388$\pm$0.015 \\
\multirow{3}*{COLLAB} &	Info-LESINN &	0.286$\pm$0.048&	0.272$\pm$0.038&	0.264$\pm$0.032&	0.260$\pm$0.026&	0.255$\pm$0.023&	0.255$\pm$0.024&	0.275$\pm$0.029&	0.319$\pm$0.033 \\
&	WL-LESINN &	0.603$\pm$0.029&	0.587$\pm$0.026&	0.535$\pm$0.029&	0.450$\pm$0.030&	 0.373$\pm$0.025 &	0.365$\pm$0.019&	0.445$\pm$0.018&	0.536$\pm$0.014 \\
&	PK-LESINN &	0.621$\pm$0.037&	0.606$\pm$0.031&	 0.558$\pm$0.046&	0.474$\pm$0.057&	 0.394$\pm$0.054&	0.382$\pm$0.051&	0.472$\pm$0.051&	0.550$\pm$0.043 \\
\multirow{3}*{hERG} &	Info-LESINN &	0.574$\pm$0.044&	0.601$\pm$0.046&	0.610$\pm$0.050&	0.628$\pm$0.053&	0.641$\pm$0.052&	0.659$\pm$0.051&	0.685$\pm$0.049&	0.701$\pm$0.048 \\
&	WL-LESINN &	0.742$\pm$0.035&	0.753$\pm$0.026&	0.764$\pm$0.027&	0.772$\pm$0.031&	0.782$\pm$0.035&	0.795$\pm$0.043&	0.802$\pm$0.048&	0.802$\pm$0.047 \\
&	PK-LESINN &	0.762$\pm$0.036&	0.769$\pm$0.037&	0.775$\pm$0.039&	0.779$\pm$0.042&	0.791$\pm$0.043&	 0.798$\pm$0.049&	0.800$\pm$0.054&	0.798$\pm$0.052\\
\hline
\end{tabular}
}
\end{table*}

\section{Training and Test Time}\label{subsectime}
Table~\ref{timetable} is the training and test time of each method on 3 datasets: REDDIT, p53 and COLLAB.
REDDIT and COLLAB are the datasets with the largest average number of nodes and edges in all 16 datasets, respectively. The dataset p53 contains the largest number of graphs. The training time of the two-stage methods considers only the graph representations/embeddings learning time. 

\begin{figure}[t!]
    \centering
    \setlength{\abovecaptionskip}{0.cm}
    \includegraphics[width=7cm]{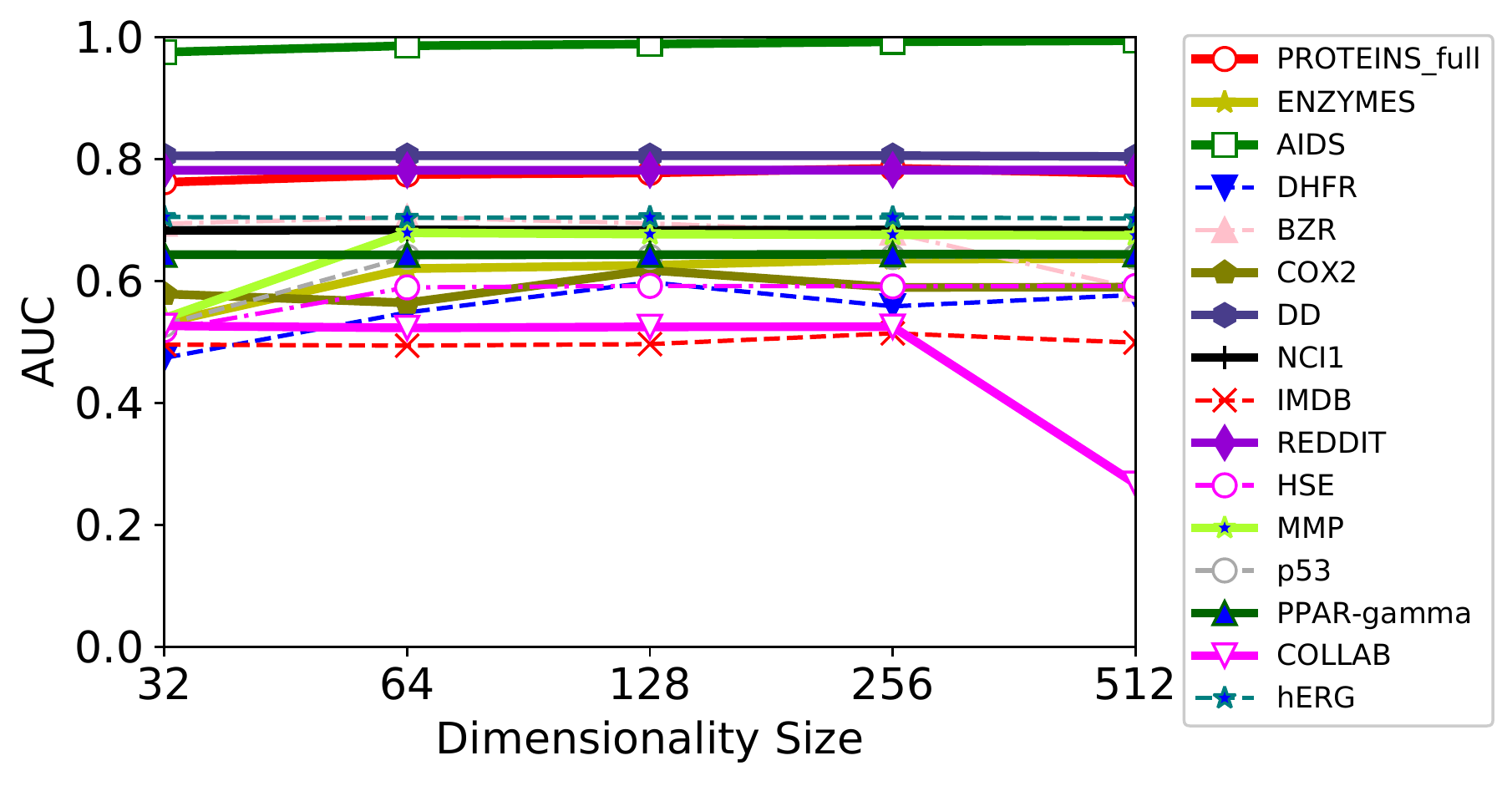}
    \caption{AUC results w.r.t. representation dimensionality.}
    \label{dimen}
\end{figure}

\section{Sensitivity Test}\label{subsec:sensitivitytestAPP}
\begin{figure}[b!]
    \centering
    \setlength{\abovecaptionskip}{0.cm}
    \includegraphics[width=7cm]{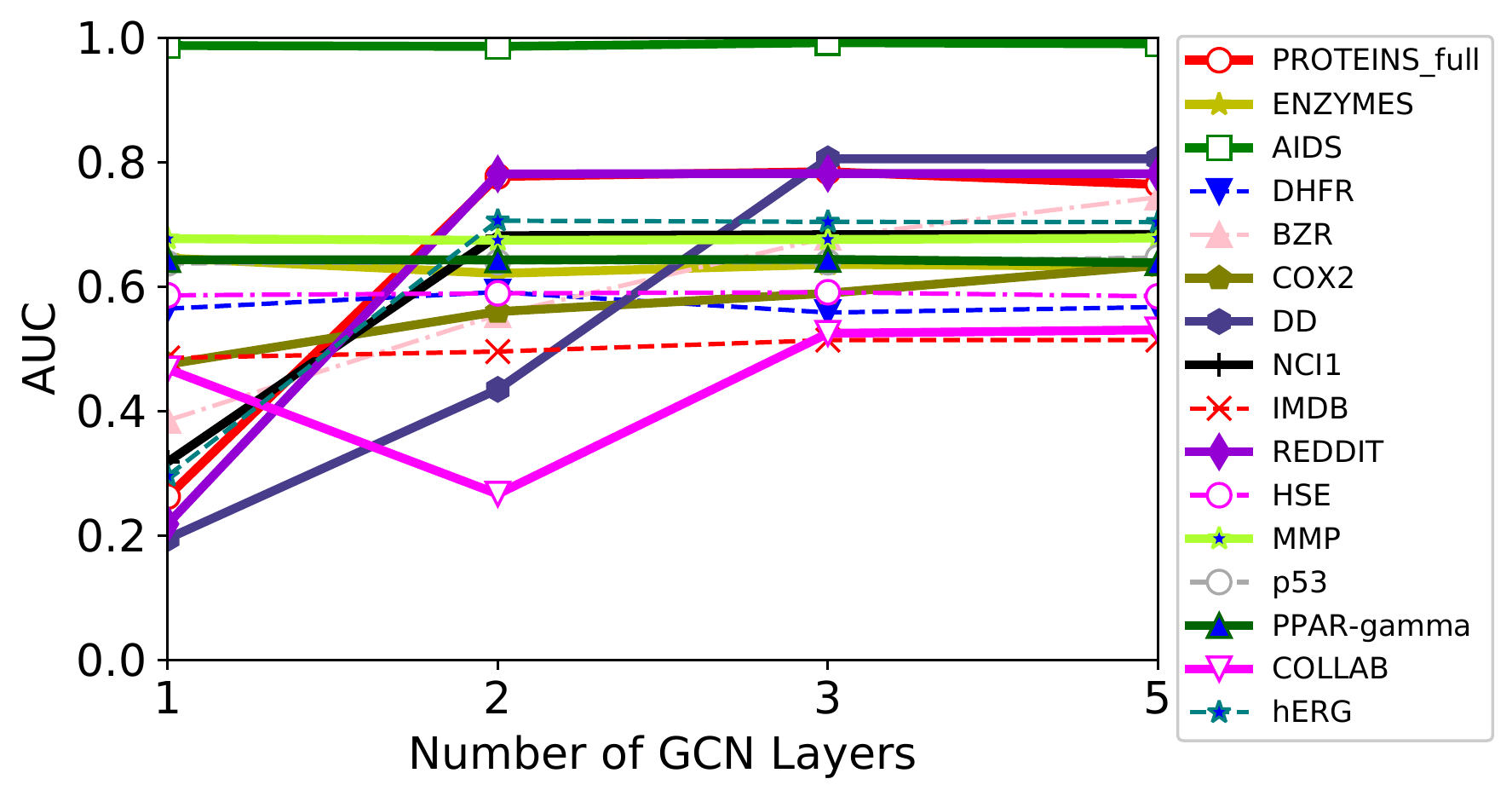}
    \caption{AUC of GLocalKD with different GCN depths.}
    \label{layer}
\end{figure}

Figure~\ref{dimen} and Figure~\ref{layer} show the performance of GLocalKD w.r.t. the GCN's output dimensionality and depth, respectively. 

\section{Detailed Results of LESINN}\label{LESINN}
Table~\ref{lesinn} shows the effect of subsample size on the performance of LESINN. We fix the ensemble size to 256 and vary the subsampling size in $\{2,4,8,16,32,64,128,256\}$ to obtain the results.

\end{document}